\PassOptionsToPackage{margin=1in}{geometry}
\documentclass{article}
\usepackage{amssymb}

\usepackage[final]{corl_2025} % Uncomment for the camera-ready ``final'' version.

\usepackage{graphics} % for pdf, bitmapped graphics files
\usepackage{amsmath} % assumes amsmath package installed
\usepackage{amssymb}  % assumes amsmath package installed
\usepackage{booktabs}
\usepackage{multirow}
\usepackage{cancel}
\usepackage{makecell}
\usepackage{graphicx}
\usepackage{booktabs}
\usepackage[table,xcdraw]{xcolor}
\usepackage[table,xcdraw]{xcolor}
\usepackage[normalem]{ulem}
\usepackage{subcaption}  
\usepackage{enumitem}  
\usepackage{pifont}
\usepackage{caption}
\usepackage{wrapfig} 
\usepackage{pgfplots}
\pgfplotsset{compat=newest}
\usepackage{tikz}
\usepackage{siunitx}
\usepackage{adjustbox}
\usepackage[margin=1in]{geometry}

\useunder{\uline}{\ul}{}

\title{LaVA-Man: Learning Visual Action Representations for Robot Manipulation}

\author{
Chaoran Zhu$^{1}$ \quad Hengyi Wang$^{2}$ \quad Yik Lung Pang$^{1}$ \quad Changjae Oh$^{1}$\\[2ex]
$^1$ Queen Mary University of London  \quad
$^2$ University College London
\\ [2ex]
{\tt \href{https://qm-ipalab.github.io/LaVA-Man}{https://qm-ipalab.github.io/LaVA-Man}}
}

\begin{document}
\makeatletter
\let\@oldmaketitle\@maketitle%
\renewcommand{\@maketitle}{\@oldmaketitle%
  \begin{center}
  \captionsetup{type=figure}
  \vspace{-15pt}
  \includegraphics[width=\textwidth]{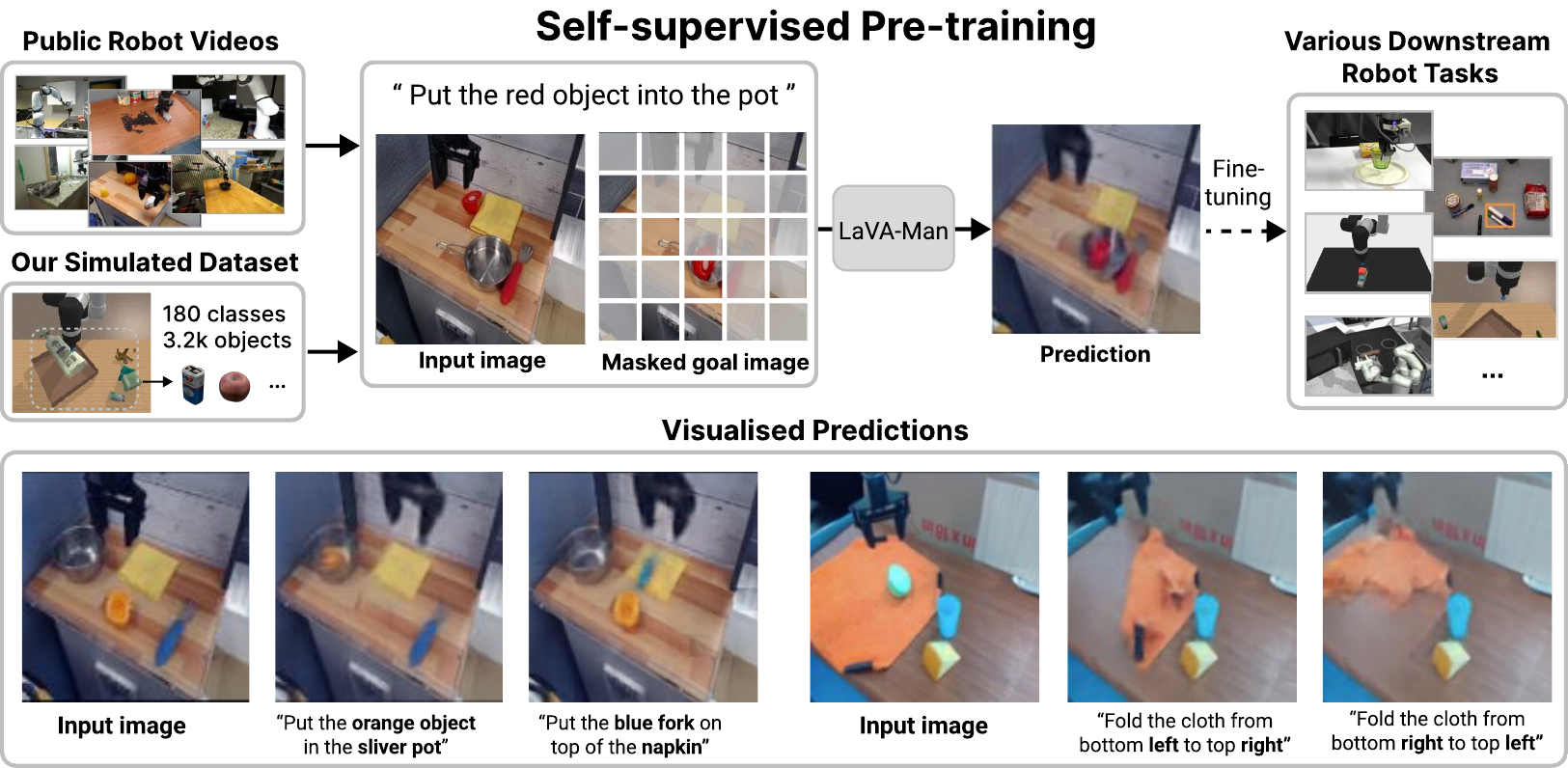}
    \caption{We introduce \textbf{LaVA-Man}, a self-supervised framework for learning visual-action representations for robot manipulation via goal image prediction. We also introduce the Omni-Object Pick-and-Place dataset to ensure the model learns a diverse, open-vocabulary-based object prior. The learned representations can be adapted to various downstream robotic perception and manipulation tasks. The samples of the goal image prediction illustrate that the learned representation can capture the underlying causality of visual state transitions in language-guided manipulation.}
    \label{fig:teaser}
    \end{center}
}
\makeatother
\maketitle

%=========================================================================

%
% --- inline annotations
%
\newcommand{\red}[1]{{\color{red}#1}}
\newcommand{\todo}[1]{{\color{red}#1}}
\newcommand{\TODO}[1]{\textbf{\color{red}[TODO: #1]}}
\newcommand{\y}[0]{\ding{52}}
\newcommand{\n}[0]{\ding{56}}

\definecolor{plotblue}{RGB}{163, 206, 241}
\definecolor{plotgray}{RGB}{173, 181, 189}

% --- disable by uncommenting  
% \renewcommand{\TODO}[1]{}
% \renewcommand{\todo}[1]{#1}

% \newcommand\hcancel[2][black]{\setbox0=\hbox{$#2$}%
% \rlap{\raisebox{.5\ht0}{\textcolor{#1}{\rule{\wd0}{0.5pt}}}}#2}
\newcommand\hcancel[2][black]{\setbox0=\hbox{$#2$}%
\rlap{\raisebox{.5\ht0}{\textcolor{#1}{\rule{\dimexpr\wd0+0.25em}{0.5pt}}}}#2}

\newcommand{\vect}[1]{\boldsymbol{\mathbf{#1}}}
\newcommand{\norm}[1]{\left\lVert#1\right\rVert}

\def\rvepsilon{{\mathbf{\epsilon}}}
\def\rvtheta{{\mathbf{\theta}}}
\def\rvlambda{{\mathbf{\lambda}}}
\def\rva{{\mathbf{a}}}
\def\rvb{{\mathbf{b}}}
\def\rvc{{\mathbf{c}}}
\def\rvd{{\mathbf{d}}}
\def\rve{{\mathbf{e}}}
\def\rvf{{\mathbf{f}}}
\def\rvg{{\mathbf{g}}}
\def\rvh{{\mathbf{h}}}
\def\rvu{{\mathbf{i}}}
\def\rvj{{\mathbf{j}}}
\def\rvk{{\mathbf{k}}}
\def\rvl{{\mathbf{l}}}
\def\rvm{{\mathbf{m}}}
\def\rvn{{\mathbf{n}}}
\def\rvo{{\mathbf{o}}}
\def\rvp{{\mathbf{p}}}
\def\rvq{{\mathbf{q}}}
\def\rvr{{\mathbf{r}}}
\def\rvs{{\mathbf{s}}}
\def\rvt{{\mathbf{t}}}
\def\rvu{{\mathbf{u}}}
\def\rvv{{\mathbf{v}}}
\def\rvw{{\mathbf{w}}}
\def\rvx{{\mathbf{x}}}
\def\rvy{{\mathbf{y}}}
\def\rvz{{\mathbf{z}}}

\def\rmA{{\mathbf{A}}}
\def\rmB{{\mathbf{B}}}
\def\rmC{{\mathbf{C}}}
\def\rmD{{\mathbf{D}}}
\def\rmE{{\mathbf{E}}}
\def\rmF{{\mathbf{F}}}
\def\rmG{{\mathbf{G}}}
\def\rmH{{\mathbf{H}}}
\def\rmI{{\mathbf{I}}}
\def\rmJ{{\mathbf{J}}}
\def\rmK{{\mathbf{K}}}
\def\rmL{{\mathbf{L}}}
\def\rmM{{\mathbf{M}}}
\def\rmN{{\mathbf{N}}}
\def\rmO{{\mathbf{O}}}
\def\rmP{{\mathbf{P}}}
\def\rmQ{{\mathbf{Q}}}
\def\rmR{{\mathbf{R}}}
\def\rmS{{\mathbf{S}}}
\def\rmT{{\mathbf{T}}}
\def\rmU{{\mathbf{U}}}
\def\rmV{{\mathbf{V}}}
\def\rmW{{\mathbf{W}}}
\def\rmX{{\mathbf{X}}}
\def\rmY{{\mathbf{Y}}}
\def\rmZ{{\mathbf{Z}}}

\def\gA{{\mathcal{A}}}
\def\gB{{\mathcal{B}}}
\def\gC{{\mathcal{C}}}
\def\gD{{\mathcal{D}}}
\def\gE{{\mathcal{E}}}
\def\gF{{\mathcal{F}}}
\def\gG{{\mathcal{G}}}
\def\gH{{\mathcal{H}}}
\def\gI{{\mathcal{I}}}
\def\gJ{{\mathcal{J}}}
\def\gK{{\mathcal{K}}}
\def\gL{{\mathcal{L}}}
\def\gM{{\mathcal{M}}}
\def\gN{{\mathcal{N}}}
\def\gO{{\mathcal{O}}}
\def\gP{{\mathcal{P}}}
\def\gQ{{\mathcal{Q}}}
\def\gR{{\mathcal{R}}}
\def\gS{{\mathcal{S}}}
\def\gT{{\mathcal{T}}}
\def\gU{{\mathcal{U}}}
\def\gV{{\mathcal{V}}}
\def\gW{{\mathcal{W}}}
\def\gX{{\mathcal{X}}}
\def\gY{{\mathcal{Y}}}
\def\gZ{{\mathcal{Z}}}

\def\sA{{\mathbb{A}}}
\def\sB{{\mathbb{B}}}
\def\sC{{\mathbb{C}}}
\def\sD{{\mathbb{D}}}
\def\sE{{\mathbb{E}}}
\def\sF{{\mathbb{F}}}
\def\sG{{\mathbb{G}}}
\def\sH{{\mathbb{H}}}
\def\sI{{\mathbb{I}}}
\def\sJ{{\mathbb{J}}}
\def\sK{{\mathbb{K}}}
\def\sL{{\mathbb{L}}}
\def\sM{{\mathbb{M}}}
\def\sN{{\mathbb{N}}}
\def\sO{{\mathbb{O}}}
\def\sP{{\mathbb{P}}}
\def\sQ{{\mathbb{Q}}}
\def\sR{{\mathbb{R}}}
\def\sS{{\mathbb{S}}}
\def\sT{{\mathbb{T}}}
\def\sU{{\mathbb{U}}}
\def\sV{{\mathbb{V}}}
\def\sW{{\mathbb{W}}}
\def\sX{{\mathbb{X}}}
\def\sY{{\mathbb{Y}}}
\def\sZ{{\mathbb{Z}}}

\newcommand{\boldparagraph}[1]{\vspace{0.1cm}\noindent{\bf #1}}

\newcommand{\boldparagraphnovspace}[1]{\vspace{-0.00cm}\noindent\textbf{#1}}
\begin{abstract}
Visual-textual understanding is essential for language-guided robot manipulation. Recent works leverage pre-trained vision-language models to measure the similarity between encoded visual observations and textual instructions, and then train a model to map this similarity to robot actions.
However, this two-step approach limits the model to capture the relationship between visual observations and textual instructions, leading to reduced precision in manipulation tasks.
We propose to learn visual-textual associations through a self-supervised pretext task: reconstructing a masked goal image conditioned on an input image and textual instructions. This formulation allows the model to learn visual-action representations without robot action supervision. The learned representations can then be fine-tuned for manipulation tasks with only a few demonstrations.
We also introduce the \textit{Omni-Object Pick-and-Place} dataset, which consists of annotated robot tabletop manipulation episodes, including 180 object classes and 3,200 instances with corresponding textual instructions. This dataset enables the model to acquire diverse object priors and allows for a more comprehensive evaluation of its generalisation capability across object instances.
Experimental results on the five benchmarks, including both simulated and real-robot validations, demonstrate that our method outperforms prior art.
\end{abstract}

\keywords{Robot manipulation, self-supervised representation learning.} 

\section{Introduction}

Language-guided robot manipulation is a fundamental task in robotics, enabling embodied agents to interpret human instructions and interact with complex environments. This task requires learning a robust representation that effectively associates visual observations with textual instructions, and can be readily mapped to the corresponding robot actions.

The key challenge is to learn such representations in a scalable manner without heavily relying on robot action annotations, such as ground-truth affordance or joint angles. Several works~\cite{shridhar2022cliport,wangprogrammatically,shridhar2023perceiver,gem} leverage pre-trained vision-language foundation models, such as CLIP~\cite{radfordLearning2021a}, which encode images and text into a unified embedding space to serve as the visual-textual representations. These methods compute the cosine similarity between image and text embeddings and learn to map this similarity to robot actions. However, their representations lack causal grounding as they do not capture how the input visual state and textual instructions lead to the resulting visual state after the robot executes an action. Namely, they do not learn true language-guided visual-action representations necessary for manipulation.

To address this limitation, we propose \textbf{LaVA-Man}, a self-supervised learning approach for learning \textbf{La}nguage-guided \textbf{V}isual-\textbf{A}ction representations for robot \textbf{Man}ipulation. Given textual instructions and visual observations before and after a manipulation, we mask the goal image and train the model to predict its masked content, conditioned on the input image and the language instruction, with minimal guidance from the unmasked regions.  Unlike prior approaches that adopt pretext tasks for general-purpose vision (e.g., masked image reconstruction~\cite{ho2022video,karamchetiLanguageDriven2023}), our goal-image prediction objective captures the underlying causality of manipulation: it enables the model to implicitly learn the association between visual dynamics and action semantics, which is critical for robotic reasoning.

To learn representations that capture diverse and open-vocabulary textual instructions, it is important to train on sequences involving various object instances. However, existing manipulation datasets, such as Ravens~\cite{zeng2021transporter} and VIMA~\cite{jiangVIMA2023}, suffer from limited object diversity. To this end, we introduce the \textit{Omni-Object Pick-and-Place} (\textit{OOPP}) dataset, a simulated tabletop manipulation dataset that consists of 180 object classes and 3,200 unique instances. Built on high-quality real-scanned meshes with language annotations from OmniObject3D~\cite{omniobj}, OOPP includes curated, scaled objects suitable for manipulation, with scene sequences automatically simulated using PyBullet Gym~\cite{benelot2018}. For a comprehensive evaluation, we hold out 20 object classes as unseen for inter-class generalisation evaluation, and reserve a subset of instances from 20 seen categories for intra-class evaluation.

We train LaVA-Man on a mixture of synthetic data from the proposed OOPP dataset and real-world robot videos from Bridge~\cite{walke2023bridgedata} and DROID~\cite{droid}. Once pre-trained, our model can be efficiently fine-tuned with only a few demonstrations for various downstream robotic perception and manipulation tasks. Our contributions are summarised as:

\begin{itemize}[itemsep=0pt, parsep=0pt, topsep=0pt, partopsep=0pt, leftmargin=*]
    \item We propose a self-supervised approach for learning a robust, versatile visual-action representation that can be efficiently fine-tuned on various robotic tasks with a few demonstrations.
    \item We introduce a new dataset based on the existing pick-and-place benchmarks~\cite{shridhar2022cliport,zeng2021transporter, jiangVIMA2023}. Our dataset features 3,200 unique, real-scanned objects from 180 categories.
    \item We validate our approach on five downstream robotic tasks, including simulated and real-world environments, and establish state-of-the-art performance.
\end{itemize}

\section{Related work}
\boldparagraph{Vision-based robot manipulation.}
Classic vision-based manipulation methods usually rely on a two-stage pipeline, where vision-based perception~\cite{xiang2017posecnn,zhu2014single,zeng2017multi,deng2020self} is utilised first, followed by a control algorithm~\cite{du2021vision}. Recent methods have shifted towards end-to-end learning frameworks~\cite{zakka2020form2fit, song2020grasping, wu2019learning, wu2022transporters}. However, these methods often rely on limited training labels and demonstrate limited generalisation ability. 
CLIPort~\cite{shridhar2022cliport} and its variants~\cite{wangprogrammatically, gem} show enhanced generalisation to open-set problems by leveraging the semantic understanding provided by CLIP~\cite{radfordLearning2021a}. However, their CLIP-based representations simply provide visual-textual similarity without causal grounding and limit their capabilities in capturing the underlying causality of the language-guided manipulation. Besides CLIP, other recent methods~\cite{a3vlm, omnimanip, manipllm} utilise large-scale vision-language models (VLMs) to achieve generalisation. But the reliance on large-scale models may introduce practical limitations.

\boldparagraph{Self-supervised visual pre-training.}
Self-supervised visual pre-training has become a fundamental approach for learning generalisable visual representations from large-scale, unlabeled data through various pretext tasks such as solving jigsaw puzzles~\cite{doersch2015unsupervised}, image colourisation~\cite{zhang2016colorful}, rotation estimation~\cite{gidaris2018unsupervised}, inpainting~\cite{pathak2016context}, and instance discrimination~\cite{he2020momentum,chen2021exploring}. Inspired by masked token prediction in BERT~\cite{kenton2019bert}, Masked-autoencoders (MAE)~\cite{heMasked2022} mask portions of input images and train the model to reconstruct the missing patches. This method shows great success in learning generalisable representations and leads to various follow-up works on learning temporal correspondence~\cite{guptaSiamese2023, tokenB} and spatial information~\cite{weinzaepfel2022croco}. We propose goal-image prediction as a pretext task that implicitly learns the causality in language-guided manipulation, enabling our model to learn visual-action representations that associate visual states with robot actions conditioned on instructions.

\boldparagraph{Visual pre-training for robotics.}
Visual pre-training can be applied to robotic manipulation to enhance the generalisation ability~\cite{karamchetiLanguageDriven2023, yen2020learning}. They rely on various pre-training methods, such as contrastive learning~\cite{kostrikov2020image,laskin2020reinforcement}, MAE~\cite{karamchetiLanguageDriven2023, mvp1, mvp2}, and other perception tasks~\cite{mpi, chenSUGAR2024}. For instance, the approach in~\cite{yen2020learning} pre-trains a visual backbone using classic tasks such as image classification and object detection. VIP~\cite{vip}, R3M~\cite{r3m}, MVP~\cite{mvp1, mvp2}, and MCR~\cite{mcr} aim to learn implicit features for robotic tasks by training on video data, while Thiea~\cite{theia} and SUSIE~\cite{susie} try to distil knowledge from pre-trained vision foundation models. SUGAR~\cite{chenSUGAR2024} and 3D-MVP~\cite{3dmvp} extend the pre-training tasks to the 3D domain. To further improve the semantic understanding, Voltron~\cite{karamchetiLanguageDriven2023} and MPI~\cite{mpi} combine video-based tasks with text conditioning, enabling generalisation to diverse object manipulation tasks. However, video data often contains temporal redundancy, which may allow models to exploit smooth temporal transitions rather than learning meaningful representations~\cite{wangMasked2023}. 
On the other hand, we design a new pretext paradigm to directly predict goal image with textual instructions via asymmetric masking~\cite{guptaSiamese2023,weinzaepfel2022croco}.

\section{Method}

Given a visual observation $\rvo_s$ and associated language instruction $\rvl_{s\to f}$, our goal is to learn a policy that predicts a robot action $\rva_{s\to f}$ that leads to a goal observation $\rvo_f$. We formulate this problem into two stages: \textit{1) Visual-action representation learning}: We first train a model $f_\theta$ that outputs the predicted goal image, $\hat{\rvo}_f$, from $\rvo_s$ and $\rvl_{s\to f}$. This self-supervised task encourages the model to align visual and textual modalities, capturing how the instruction transforms the scene. \textit{2) Robot action prediction}: We then fine-tune the model by attaching an action prediction head that maps the learned representation to a robot action $\hat{\rva}_{s\to f}$. This action may take various forms depending on the task setting, e.g., $\hat{\rva}_{s\to f} \in SE(2)$ for tabletop manipulation or $\hat{\rva}_{s\to f} \in \mathbb{R}^9$ for joint angle control.

\begin{figure}[t]
  \centering
  \includegraphics[width=\linewidth]{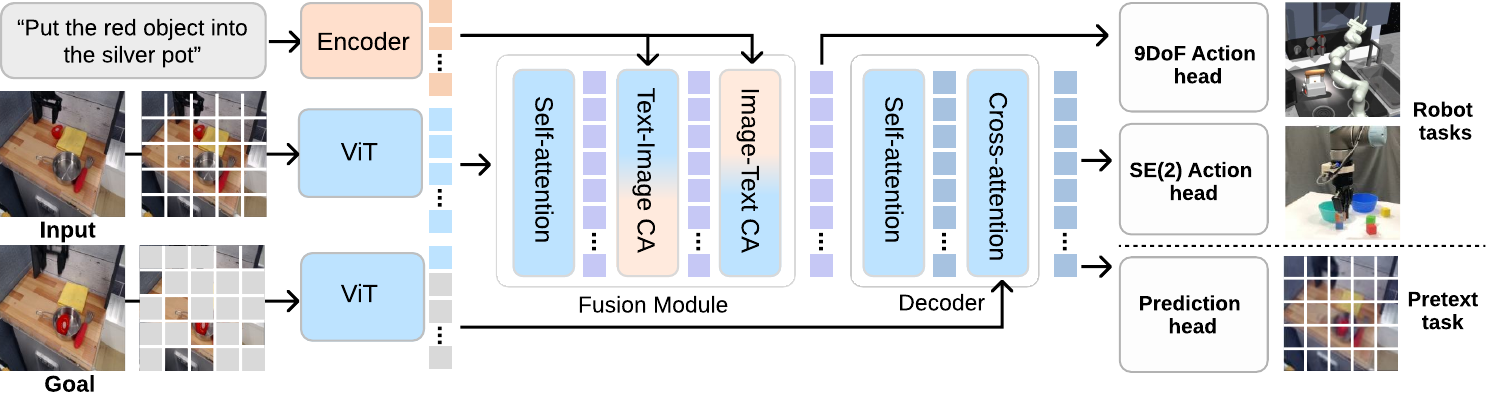}
  \caption{\textbf{Network structure.} We use a fixed backbone while adapting different output heads for pretext and downstream robot tasks. We use a Siamese ViT encoder with asymmetric masking applied to the goal image only. Visual features from the input are first fused with text features and then integrated with those extracted from the masked goal image in the decoder. During inference, the goal image is fully masked as it is unknown. Additional output heads can be incorporated to support a wider range of downstream tasks. KEY – CA: cross-attention.}
  \label{fig:ab_probe}
\end{figure}

\subsection{Learning visual-action representations}\label{sec-method-lava}

We propose a self-supervised approach of asymmetric masking to learn visual-action representations using the pretext task, which aims to reconstruct a partially masked goal image given $\rvo_s$ and $\rvl_{s\to f}$. Fig.~\ref{fig:ab_probe} shows LaVA-Man's architecture and the proposed pretext task of goal image prediction.  

\boldparagraph{Vision encoder.} Let $\rvo_s$ and $\rvo_f$ be two images that capture the scene before and after the robot manipulation. Both images are firstly divided into $N$ non-overlapping patches, where some patches in $\rvo_f$ are randomly masked, transforming $\rvo_f$ into $\tilde{\rvo}_f$. We design a pretext task that reconstructs ${\rvo}_f$ based on ${\rvo}_s$, $\tilde{\rvo}_f$, and a text embedding $\rve_{s\to f}$ obtained from $\rvl_{s\to f}$:
\begin{equation}
\label{eq:eq1}
\begin{aligned}
    \rvh_{s\to f} &= \Phi(\rvo_s, \tilde{\rvo}_f, \rve_{s\to f}),\\
    \hat{\rvo}_f &= \Psi_p(\rvh_{s\to f}).
\end{aligned}
\end{equation}

\noindent 
Here $\Phi$ consists of a siamese Vision Transformer (ViT) encoder that encodes the patches in $\rvo_s$ and $\rvo_f$ into the features $\rvv_s, \rvv_f$ and a lightweight decoder that decodes $\rvv_s$, $\rvv_f,$ and $\rve_{s\to f}$ to a feature vector $\rvh_{s\to f}$. $\Psi_p$ is the goal-image prediction head that outputs the reconstruction $\hat{\rvo}_f$ given $\rvh_{s\to f}$. We use the CLIP text encoder to encode the text embeddings $\rve_{s\rightarrow f}$ from $\rvl_{s\rightarrow f}$.

\boldparagraph{Visual-textual fusion.} Inspired by GLIP~\cite{glip}, we perform a multi-stage cross-attention to fuse information from different modalities. First, we use text-to-image and image-to-text attention: $\mathrm{cross\_attn}(\rvv_s, \rve_{s\to f})$ and $\mathrm{cross\_attn}(\rve_{s\to f}, \rvv_s)$, to fuse the initial visual state and language.  Then, $\rvv_f$ queries this fused feature via $\mathrm{cross\_attn}(\rvv_s, \rvv_f)$ to produce $\rvh_{s\to f}$. This enables the model to condition on both the initial image and the language instruction while grounding the goal image.

\boldparagraph{Goal-image prediction head.} 
The per-patch feature $\rvh_{s\to f}$ is passed to a lightweight MLP head $\Psi_p$ to generate the per-pixel RGB values for each patch in the predicted goal image $\hat{\rvo}_f$. We supervise the model using an $\gL_2$ loss between $\hat{\rvo}_f$ and the ground-truth $\rvo_f$.

\subsection{Robot action prediction}\label{sec-method-action-prediction}
\label{sec:robot_action_prediction}
Following the previous stage, we fine-tune our model for downstream robot manipulation tasks with an additional lightweight head $\Psi_a$ that predicts the robot action $\rva_{s\to f}$:
\begin{equation}
    \rva_{s\to f} = \Psi_a(\rvh_{s\to f}, \rvo_s).
    \label{eq:action1}
\end{equation}

\noindent
At test time, the model does not have $\rvo_f$ and $\tilde{\rvo}_f$. However, our pretext task is designed to learn from $\tilde{\rvo}_f$ with a high mask ratio, and hence our model can still predict $\rvh_{s\to f}$ as in Eq.~\ref{eq:eq1} while using fully masked $\tilde{\rvo}_f$.
We therefore include $\hat{\rvo}_f$ as additional input to the action head $\Psi_a$ and rewrite Eq.~\ref{eq:action1} as:
\begin{equation}
    \rva_{s\to f} = \Psi_a(\rvh_{s\to f}, \rvo_s, \hat{\rvo}_f).
\end{equation}

\noindent

For tabletop manipulation tasks, we define the output action as $\rva_{s\to f}=(\mathcal{T}_{s}, \mathcal{T}_{f})$, where $\mathcal{T}_{s}, \mathcal{T}_{f} \in \mathbf{SE}(2)$ denote the pick and place poses of the end-effector. The model predicts an affordance map as an intermediate representation following~\cite{ shridhar2022cliport,zeng2021transporter}, from which $\mathcal{T}_{s}$ and $\mathcal{T}_{f}$ are extracted via a softmax operation. For the task of predicting joint angles on specific robot arms, we drop the decoder as ~\cite{karamchetiLanguageDriven2023, mpi} and represent each action as a 9-DoF vector $\rva_{s \to f} \in \mathbb{R}^9$, comprising seven joint angles and two indicators for grasp status.

\section{The Omni-Object Pick-and-Place dataset}

\begin{figure}[t]
\centering
\begin{tabular}{@{}c@{\hspace{5pt}}c@{}}
    \raisebox{50pt}{%
        \resizebox{0.495\textwidth}{!}{%
            \begin{tabular}{@{}lcccccc@{}}
            \toprule
            \multicolumn{1}{l}{Dataset} & \begin{tabular}[c]{@{}c@{}}Num of \\ classes\end{tabular} & \begin{tabular}[c]{@{}c@{}}Num of \\ instances\end{tabular} & \begin{tabular}[c]{@{}c@{}}Inter-class \\ variation\end{tabular} & \begin{tabular}[c]{@{}c@{}}Intra-class \\ variation\end{tabular} & \begin{tabular}[c]{@{}c@{}}Real scanned \\ objects\end{tabular} \\
            \midrule
            RLBench$^\dag$~\cite{rlbench} & 100 & 100 & \y & \n & \y \\
            LIBERO$^\dag$~\cite{libero} & 100 & 100 & \y & \n & \y \\
            Ravens~\cite{shridhar2022cliport} & 52 & 52 & \y & \n & \y \\
            VIMA~\cite{jiangVIMA2023} & 20 & 1800$^\ast$ & \y & \y & \n \\
            \rowcolor[HTML]{ECF4FF} \textbf{OOPP (ours)} & 180 & 3200 & \y & \y & \y \\
            \rowcolor[HTML]{FFFFFF}
            \bottomrule
            \addlinespace[\belowrulesep]
            \multicolumn{6}{l}{\parbox{0.8\linewidth}{\footnotesize{$^{\dag}$RLBench and LIBERO focus on complex manipulation actions rather than diverse object interactions. We approximate the object number by the number of tasks as there is little intra-class object variation.}}} \\
            \addlinespace[\belowrulesep]
            \multicolumn{6}{l}{\parbox{0.8\linewidth}{\footnotesize{$^{\ast}$VIMA provides 90 different textures per object. We treat each textured variant as a distinct instance here.}}}
            \end{tabular}%
        }%
    }
    &
    \includegraphics[width=0.495\textwidth]{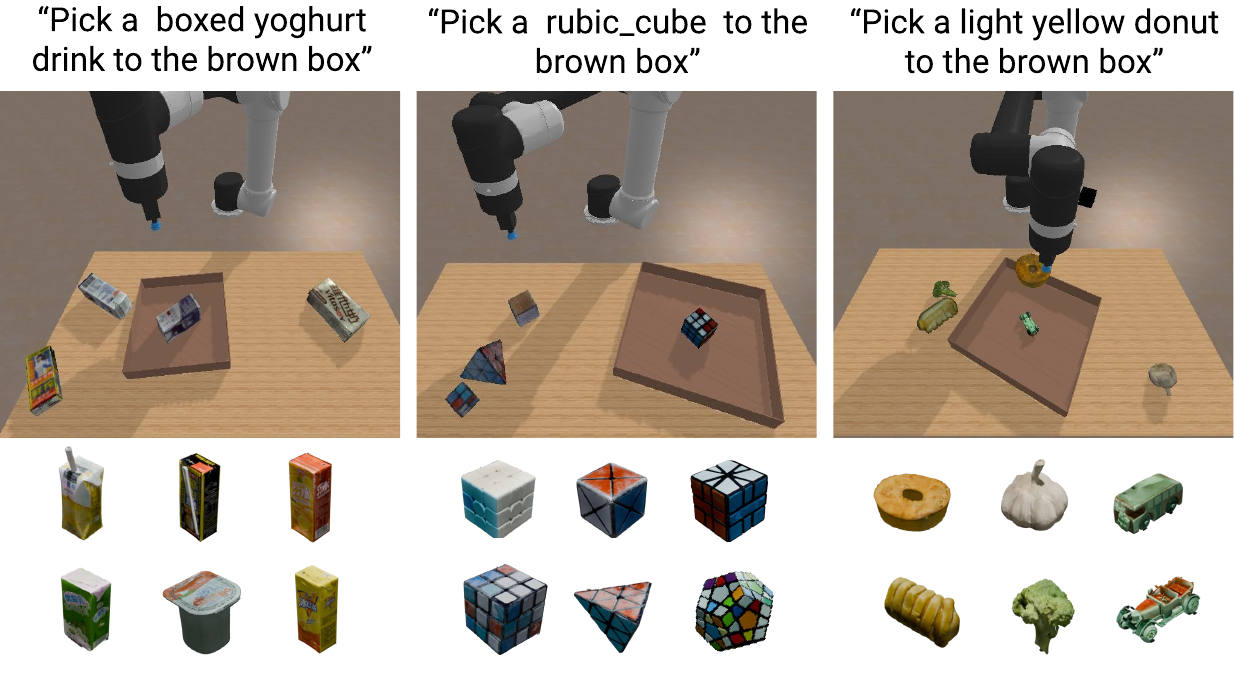} \\
    {\footnotesize (a) Quantitative comparison} & {\footnotesize (b) Visualised examples }
\end{tabular}
\caption{We show (a) Quantitative statistics of the proposed OOPP dataset compared to several existing datasets. (b) Visualised examples of manipulation sequences in OOPP datasets. The first row shows objects seen by the models and the second row shows unseen objects.}
\label{fig:oopp}
\end{figure}

%\input{figures/dataset_vis}
%\input{figures/object_mesh}

 % \boldparagraph{Motivation} 
% We present \textit{Omni-Object Pick-and-Place} (OOPP) dataset together with our method. 
%  Existing tabletop simulations such as Ravens~\citep{zeng2021transporter} and VIMA~\cite{jiangVIMA2023} suffer from limited object diversity, making it difficult to assess a model’s ability to generalise to unseen objects in zero-shot scenarios. In addition, their language instructions are largely template-based, which restricts evaluation to fixed, narrow vocabularies and fails to test a model’s capacity to understand open-vocabulary natural language. Other widely used simulators, including Franka Kitchen~\citep{franka_kitchen}, RLBench~\citep{rlbench}, and LEBRO~\citep{libero}, focus primarily on task diversity while similarly overlooking the object and language variation. Consequently, there is currently no simulation benchmark capable of evaluating a model's ability to ground diverse, open-vocabulary natural language instructions to visual observations involving both inter-class (across different categories) and intra-class (within-category variations)—and to perform corresponding robotic manipulation tasks. In addition, with the help of OOPP dataset, we can generate a large amount of data for our pre-training methods.

\boldparagraph{Overview.} 
We introduce the OOPP dataset, a tabletop simulation benchmark consisting of 3,200 unique real-scanned objects across 180 distinct categories. As shown in Fig.~\ref{fig:oopp} (a), existing tabletop simulation datasets such as Ravens~\cite{zeng2021transporter} and VIMA~\cite{jiangVIMA2023} typically exhibit limited object diversity and rely on template-based language instructions. This restricts trained models to fixed, narrow vocabularies and prevents them from generalising to open-vocabulary settings. In contrast, OOPP consists of diverse object instances paired with rich language descriptions, enabling models to learn a broad range of object priors. OOPP also allows a comprehensive evaluation of a model’s ability to ground diverse, open-vocabulary instructions to visual observations involving both inter-class and intra-class variations.

\boldparagraph{Data annotation.}
OOPP is built upon the previous benchmarks~\citep{zeng2021transporter, jiangVIMA2023} in the Pybullet Gym environment~\citep{benelot2018}. We adopt the real-scanned objects from the OmniObject3D dataset~\citep{omniobj}, filtering out the objects unsuitable for tabletop manipulation by retaining only objects with diameters between 4cm and 40cm. During simulation, we allow the robot to perform automatic manipulation with a pre-defined placeholder for collecting action annotations. These placeholders are then replaced with real objects. To ensure spatial diversity and prevent overcrowding, we adopt a KDTree-based spatial partitioning strategy inspired by~\cite{shridhar2022cliport}, subdividing scenes into feasible regions for object placement. Based on the rich language description for each object provided by the original OmniObjects3D dataset~\citep{omniobj}, we generate diverse language instructions that are beyond just using object names, offering more natural and varied descriptions of object appearances as in Fig.~\ref{fig:oopp}.

\boldparagraph{Evaluation.}
To evaluate both intra- and inter-class generalisation, we partition the dataset into three mutually exclusive subsets. We use 160 object classes for training. For intra-class generalisation, we hold out a subset of instances from 20 categories in the training set for testing. For inter-class generalisation, we reserve 20 object categories that are entirely unseen during training. Unseen classes are sampled from four high-level semantic groups—\textit{Food}, \textit{Daily-use \& Tools}, \textit{Entertainment}, and \textit{Others}—with each group contributing 3–8 classes.
We define two manipulation tasks: 1) \textit{packing-objects-group}, where identical objects are packed simultaneously, and 2) \textit{packing-objects-sequence}, where different objects are packed sequentially. Together, these variants yield six different tasks. Please refer to the supplementary material for more details.

\begin{table*}[tb]\large
\centering
\caption{\textbf{Results on the Ravens benchmark~\cite{zeng2021transporter}.} We report the results of multi-task performance trained on 1,000 demonstrations.}
\label{tab:multi}
\resizebox{0.95\textwidth}{!}{%
\renewcommand{\arraystretch}{1.05}
\setlength{\tabcolsep}{3pt}
\begin{tabular}{cccccccccccc}
\toprule
                & \multicolumn{2}{c}{\textbf{Seen objects}} & \multicolumn{2}{c}{\textbf{Unseen objects}} & \multicolumn{6}{c}{\textbf{Coloured geometries}}                     &      \\    \cmidrule(lr){2-3}
                   \cmidrule(lr){4-5}
                   \cmidrule(lr){6-11}
\multirow{-2}{*}{\textbf{Tasks}} &
  \makecell{Packing \vspace{-1mm} \\ obj-seq} &
  \makecell{Packing \vspace{-1mm} \\ obj-grp} &
  \makecell{Packing \vspace{-1mm}\\ obj-seq} &
  \makecell{Packing \vspace{-1mm} \\ obj-grp} &
  \makecell{Put block \vspace{-1mm}\\ in bowl} &
  \makecell{Stack block \vspace{-1mm}\\ pyramid} &
  \makecell{Towers \vspace{-1mm}\\ of hanoi} &
  \makecell{Packing \vspace{-1mm}\\ boxes pairs} &
  \makecell{Assembling \vspace{-1mm}\\ kits} &
  \makecell{Separating \vspace{-1mm}\\ piles} &
  \multirow{-2}{*}{\textbf{Average}} \\ 
  \midrule
\rowcolor[HTML]{EFEFEF} 
\multicolumn{12}{l}{\textit{Pick-and-place methods}}                                                                                          \\ \midrule
\multicolumn{1}{l}{Transporter~\cite{zeng2021transporter}}     & 0.49            & 0.59           & 0.25             & 0.45            & 0.23 & 0.02 & 0.1  & 0.43 & 0.20   & 0.44 & 0.32 \\
\multicolumn{1}{l}{CLIP only~\cite{radfordLearning2021a}}       & 0.57            & 0.72           & 0.49             & 0.63            & 0.43 & 0.11 & 0.2  & 0.5  & 0.36  & 0.47 & 0.45 \\
\multicolumn{1}{l}{RN50-BERT~\cite{shridhar2022cliport}}       & 0.46            & 0.64           & 0.41             & 0.59            & 0.45 & 0.02 & 0.21 & 0.47 & 0.27  & 0.5  & 0.40 \\
\multicolumn{1}{l}{CLIPort$^{\ast}$~\cite{shridhar2022cliport}}       & 0.80             & \textbf{0.89}           & 0.55             & 0.74            & 0.59 & 0.38 & 0.69 & 0.80  & 0.60 & 0.71 & 0.68 \\
\multicolumn{1}{l}{CLIPort~\cite{shridhar2022cliport}}         & 0.78            & 0.83           & 0.57             & 0.77            & 0.84 & \textbf{0.69} &\textbf{ 0.82} & 0.83 & 0.54 & 0.58 & 0.73 \\ \midrule
\rowcolor[HTML]{EFEFEF} 
\multicolumn{12}{l}{\textit{Pre-training methods}}                                                                                          \\ \midrule
\multicolumn{1}{l}{No pre-training} &0.59 &0.72&0.40&0.56&0.33&0.22&0.18&0.64 &0.55 &0.46&0.47 \\
%\multicolumn{1}{l}{MVP} &0.37	&0.59	&0.24	&0.39	&0.39	&0.17	&0.15	&0.6	&0.65	&0.07	&0.36   \\
\multicolumn{1}{l}{Voltron~\cite{karamchetiLanguageDriven2023}} &0.58	&0.7 &0.43	&0.57	&0.80	&0.36	&0.24	&0.70	&0.60	&0.43	&0.54\\
\multicolumn{1}{l}{MPI$^\ddagger$~\cite{mpi}}  &0.56	&0.75 &0.41 &0.66	&0.67	&0.22	&0.42	&0.42	&0.46	&0.46	&0.50 \\
\rowcolor[HTML]{ECF4FF} 
\multicolumn{1}{l}{\cellcolor[HTML]{ECF4FF}Ours}  &\textbf{0.83}	 &0.84  &\textbf{0.77} &\textbf{0.83}	&\textbf{0.98}	&0.57	&0.67	&\textbf{0.94}	&\textbf{0.75}	&\textbf{0.93}	&\textbf{0.81}\\ \bottomrule
\end{tabular}%
}
\vspace{1ex}
\noindent\parbox{0.92\textwidth}{\scriptsize $^{\ast}$indicates that the model was tested using checkpoints provided by the authors, while others are trained on the same downstream data.  MPI$^\ddagger$ is our reimplementation of MPI, trained in a self-supervised manner.  KEY -- obj: objects, seq: sequence, grp: group.}
\vspace{-10pt}
\end{table*}

\begin{table}[!t]
\centering
\caption{\textbf{Results on the OOPP dataset.} We report the results trained on 1,000 demonstrations and tested on 100 demonstrations in the test set.}
\label{tab:oopp}
\resizebox{0.9\textwidth}{!}{%
\renewcommand{\arraystretch}{1.05}
\begin{tabular}{cccccccc}
\toprule
\textbf{Method} &
  % \makecell{Packing \vspace{-1mm} \\ obj-seq}  
  \makecell{Packing  \\ obj-seq } &
  \makecell{Packing \\ obj-intra-seq} &
  \makecell{Packing \\ obj-inter-seq} &
  \makecell{Packing \\ obj-grp} &
  \makecell{Packing \\ obj-intra-grp } &
  \makecell{Packing \\ obj-inter-grp} &
  \textbf{Average} \\ \midrule
% \multicolumn{1}{l}{TransportNet} & \multicolumn{1}{l}{} & \multicolumn{1}{l}{} & \multicolumn{1}{l}{} & - & - & - & -          \\
CLIPort~\cite{shridhar2022cliport} & 0.62& 0.60 & 0.48 & 0.65 &0.57 & 0.60 & 58.7 \\
% MVP                              &                      &                      &                      & - & - & - & -          \\
Voltron~\cite{karamchetiLanguageDriven2023} &0.74 &0.68  &0.54& 0.80 & 0.71 & 0.72 & 69.8         \\
{MPI}$^\ddagger$~\cite{mpi}&0.69 &0.59 &0.58 & 0.73 & 0.65& 0.67 & 65.1          \\
\rowcolor[HTML]{ECF4FF} 
Ours & \textbf{0.84} &\textbf{ 0.74} &\textbf{ 0.73}   & \textbf{0.84} & \textbf{0.86} & \textbf{0.77} & \textbf{79.6} \\ \bottomrule
\end{tabular}%
}
\vspace{1ex}
\noindent\parbox{0.9\textwidth}{\scriptsize MPI$^\ddagger$ is our reimplementation of MPI, trained in a self-supervised manner. KEY -- obj: objects, seq: sequence, intra: intra-class, inter: inter-class, grp: group.}

\end{table}

\section{Experiments}

\subsection{Setup}

\boldparagraph{Pre-training datasets.} For the pretext task, we train our model using a total of 120k samples from our synthetic OOPP dataset and real-world robot video episodes from Bridge~\cite{walke2023bridgedata} and DROID~\cite{droid}. For all robot video episodes, we extract only the first and last frames, paired with language instructions. The pre-training phase takes 24 hours with 3×A100 GPUs.

\boldparagraph{Baselines.} We compare our method against two types of methods: 1) Foundation model-based methods such as CLIPort~\cite{shridhar2022cliport}, which leverage web-scale vision-language models for pick-and-place tasks; and 2) Self/weakly supervised methods, including Voltron~\cite{karamchetiLanguageDriven2023} and MPI~\cite{mpi}, which learns representations from human/robot manipulation episodes for grounding language-guided manipulation.  Since MPI pre-training also relies on supervision from object bounding boxes, we remove the detection head and make the pre-training in a fully self-supervised manner, denoted as MPI$^\ddagger$. Unless specified, we re-train every baseline using our data for fair comparison.

\boldparagraph{Downstream tasks and benchmarks}. We evaluate our model on various downstream tasks, including both simulation and real-world environments (see supplementary for details):
\begin{itemize}[itemsep=0pt, parsep=0pt, topsep=0pt, partopsep=0pt, leftmargin=*]
    \item \textbf{Ravens}~\citep{zeng2021transporter}: This dataset provides ten different tabletop rearrangement tasks. For fair comparison, we re-train all pre-training methods and fine-tune them using the same set of demonstrations.
    \item 
    \textbf{OOPP}: We also evaluate our models on our proposed dataset with diverse object classes and instances for intra- and inter-class evaluation. The training procedure follows that of Ravens.
    \item \textbf{Referring expression grounding}~\citep{karamchetiLanguageDriven2023}: This benchmark evaluates target object localisation in cluttered scenes based on language instructions, serving as a prerequisite for visuomotor control.
    \item \textbf{Franka Kitchen}~\citep{franka_kitchen}: This popular benchmark provides comparisons with other state-of-the-art methods on five visuomotor control tasks in a simulated kitchen environment.        
    \item \textbf{Real-robot experiments}: We further evaluated our model on ten different manipulation tasks using UR5 robot arms to evaluate real-world generalisation. We capture the current observation from the top-down view and generate the affordance map for action execution accordingly.

\end{itemize}

\begin{table}[!t]
  \centering
  \caption{\textbf{Results on Referring Expression Grounding~\cite{karamchetiLanguageDriven2023} and Franka Kitchen~\cite{franka_kitchen}.} Our method outperforms self-supervised methods and is comparable to leading supervised methods. The results of existing methods are quoted from~\citep{mpi} and~\citep{karamchetiLanguageDriven2023}. Minimum, Medium and Maximum in sub-table (a) denote the scenarios categorised according to the level of clutter.} 
  \vspace{5pt}

  \begin{subtable}[t]{0.36\textwidth}
    \centering
    % \caption{Referring Expression Grounding}
    %\begin{table}[]
\Large
\centering
\label{tab:franka}
\setlength{\tabcolsep}{5pt}
\resizebox{\columnwidth}{!}{%
\raisebox{60pt}{
\begin{tabular}{@{}ccccc@{}}
% \toprule
\multicolumn{5}{c}{\textbf{(a) Referring Expression Grounding}}\\
\toprule
Method     & Minimum    & Medium    & Maximum   & Total          \\ \midrule
\rowcolor[HTML]{EFEFEF} 
\multicolumn{5}{l}{\cellcolor[HTML]{EFEFEF}\textit{Supervised pre-training methods}}    \\ \midrule
MPI~\cite{mpi}        &  0.94       & \textbf{0.98}   &    0.95        &0.96   \\
SUGAR~\cite{chenSUGAR2024}      &  0.98        & 0.97         & \textbf{0.96}     & \textbf{0.97}  \\ \midrule
\rowcolor[HTML]{EFEFEF} 
\multicolumn{5}{l}{\cellcolor[HTML]{EFEFEF}\textit{Self-supervised pre-training methods}} \\ \midrule
R-R3M~\cite{r3m}      & 0.64                & 0.68              & 0.55              & 0.63           \\
MVP~\cite{mvp1}        & 0.51                & 0.52              & 0.40              & 0.49           \\
CLIP~\cite{radfordLearning2021a}       & 0.67                & 0.77              & 0.60              & 0.68           \\
Voltron~\cite{karamchetiLanguageDriven2023}    & 0.88                & \textbf{0.97}     & 0.90              & 0.91           \\
\rowcolor[HTML]{ECF4FF} 
Ours       & \textbf{0.99}       & 0.96      & \textbf{0.96}     & \textbf{0.97}  \\ \bottomrule
\end{tabular}%
% \vspace{-20pt}
}
}
%\end{table}    
    \label{tab:referring}
  \end{subtable}
  \hfill
  \begin{subtable}[t]{0.62\textwidth}
    \centering
    % \caption{Franka Kitchen}
    
\centering
\label{tab:my-table}
\resizebox{\columnwidth}{!}{%
\begin{tabular}{@{}ccccccc@{}}
% \toprule
\multicolumn{7}{c}{\textbf{(b) Franka Kitchen}}\\
\toprule
Method & Trun knob & Open door & Flip switch & \multicolumn{1}{l}{Open microwave} & \multicolumn{1}{l}{Slide door} & \multicolumn{1}{l}{Average} \\ \midrule
\rowcolor[HTML]{EFEFEF} 
\multicolumn{7}{l}{\cellcolor[HTML]{EFEFEF}\textit{Supervised pre-training methods}}      \\ \midrule
MPI~\cite{mpi}      & {\ul 89.0}   & \textbf{57.7} & {\ul 93.7}  & { 54.0}  & \textbf{100.0} & {\ul 78.9}  \\ \midrule
\rowcolor[HTML]{EFEFEF} 
\multicolumn{7}{l}{\cellcolor[HTML]{EFEFEF}\textit{Self-supervised pre-training methods}} \\ \midrule
INSUP~\cite{insup}    & 28.0         & 18.0          & 50.0        & 26.7        & 75.7  & 39.7        \\
CLIP~\cite{radfordLearning2021a}     & 26.3         & 13.0          & 41.7        & 24.7        & 86.3  & 38.4        \\
R3M~\cite{r3m}      & 53.3         & 50.7          & 86.3        & {\ul 59.3}        & 97.0  & 69.5        \\
MVP~\cite{mvp1}      & 79.0         & 48.0          & 90.7        & 41.0        & \textbf{100.0} & 71.7        \\
Voltron~\cite{karamchetiLanguageDriven2023}  & 76.0         & 45.3          & 91.0        & 41.0        & 99.3  & 70.5        \\
\rowcolor[HTML]{ECF4FF} 
Ours     & \textbf{90.0}  & {\ul 50.7}      & \textbf{94.0} & \textbf{61.3} & \textbf{100.0}   & \textbf{79.2} \\ \bottomrule
\end{tabular}%
}

%\end{table}    
    \label{tab:franka}
  \end{subtable}  
  \label{tab:combined-results}
\end{table}

\begin{figure}[t]
  \vspace{-10pt}
  \centering
  \footnotesize
  \setlength{\tabcolsep}{1pt}
  \newcommand{\plth}{0.19\linewidth} % 控制图片宽度适配五列
  \begin{tabular}{ccccc} 
    % Row 1: Images 1–5
    \includegraphics[width=\plth]{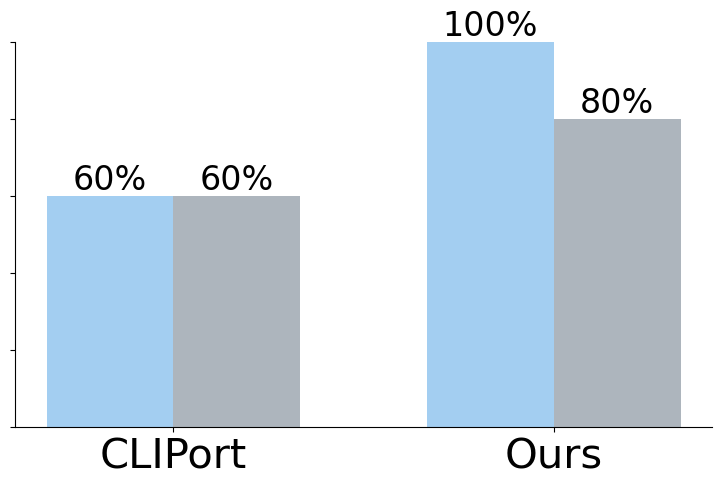} &
    \includegraphics[width=\plth]{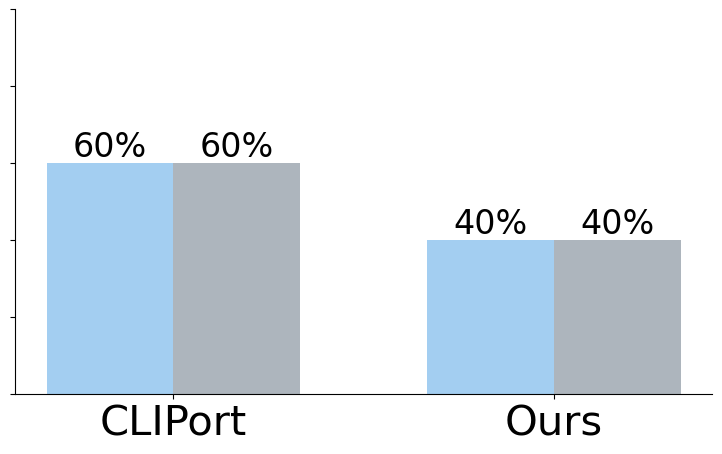} &
    \includegraphics[width=\plth]{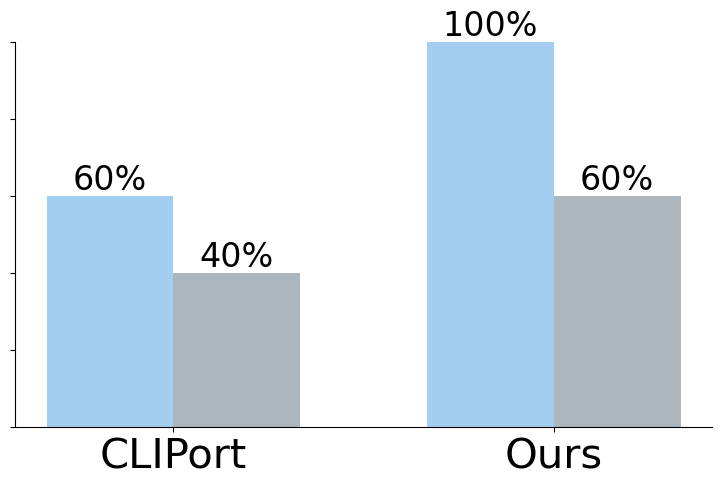} &
    \includegraphics[width=\plth]{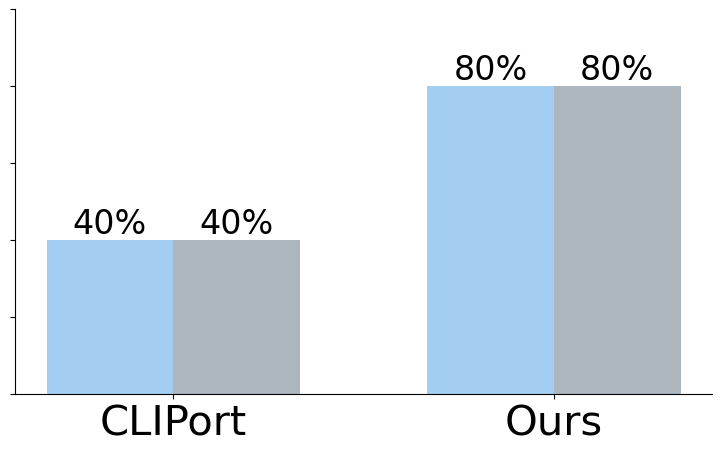} &
    \includegraphics[width=\plth]{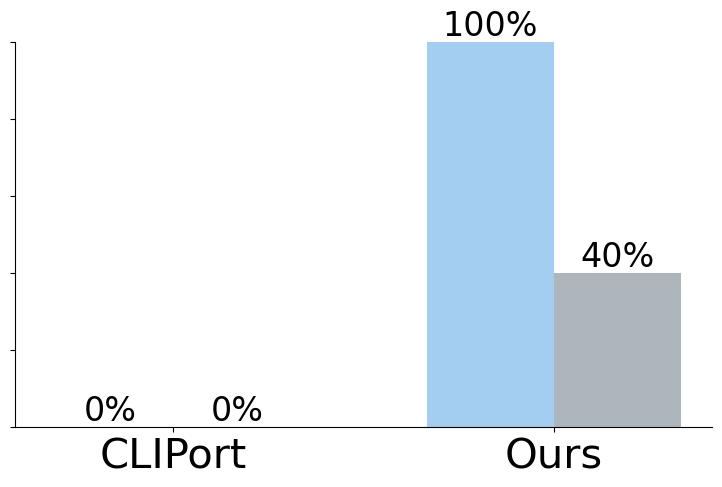} \\
   
    % Row 3: Images 6–10
    \includegraphics[width=\plth]{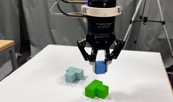} &
    \includegraphics[width=\plth]{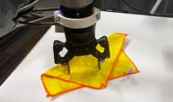} &
    \includegraphics[width=\plth]{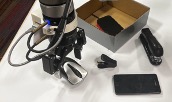} &
    \includegraphics[width=\plth]{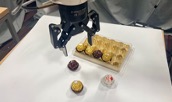} &
    \includegraphics[width=\plth]{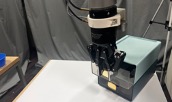} \\
    % Row 4: Captions 6–10     
     \makecell{Stacking blocks \\ {}} & \makecell{Folding cloth \\ {}} &  \makecell{Packing objects \\ {}} &\makecell{
      Packing objects \\ (\textbf{unseen objects})} & \makecell{Opening drawer \\ (\textbf{unseen task})} \\[2pt]
    
    % Row 5: Images 11–15
    \includegraphics[width=\plth]{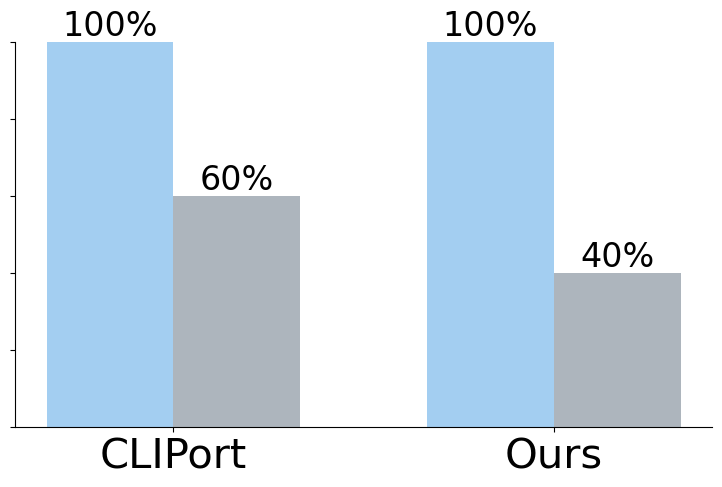} &
    \includegraphics[width=\plth]{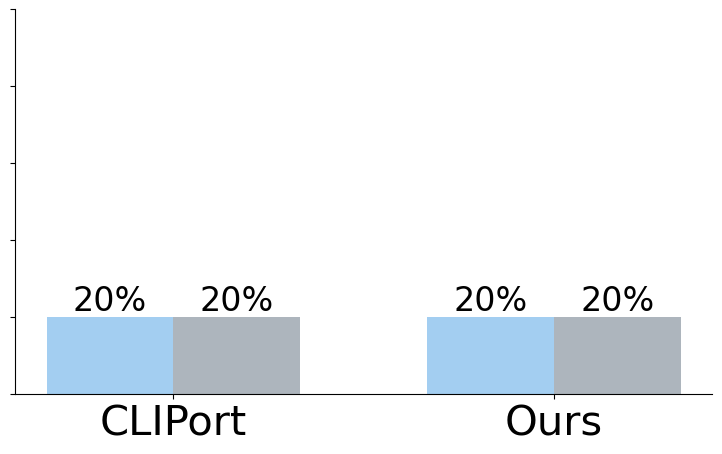} &
    \includegraphics[width=\plth]{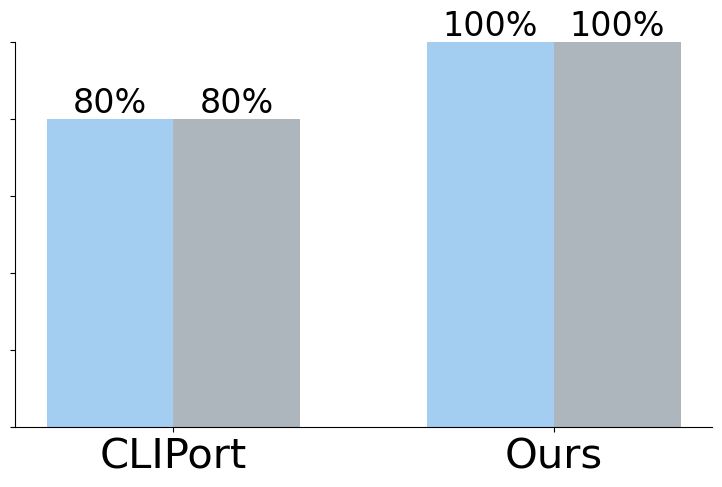} &
    \includegraphics[width=\plth]{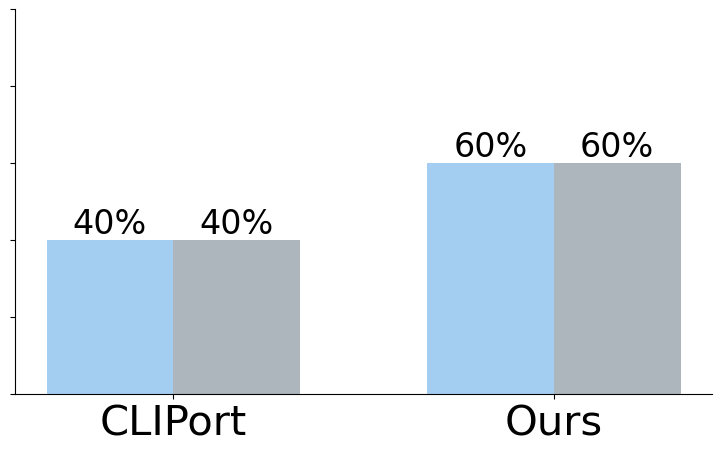} &
    \includegraphics[width=\plth]{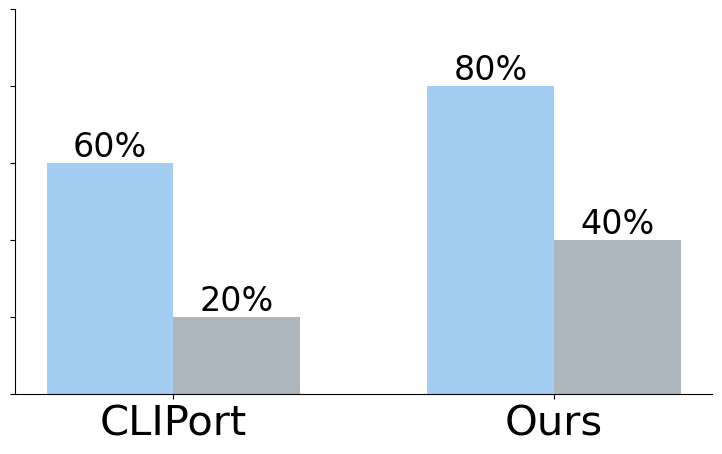} \\
    % Row 7: Images 16–20
    \includegraphics[width=\plth]{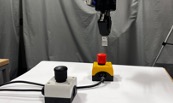} &
    \includegraphics[width=\plth]{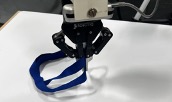} &
    \includegraphics[width=\plth]{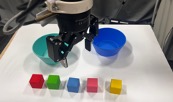} &
    \includegraphics[width=\plth]{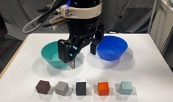} &
    \includegraphics[width=\plth]{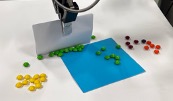} \\
    % Row 8: Captions 16–20
    \makecell{Press button\\ {}} & \makecell{Aligning rope\\ {}} & \makecell{Packing blocks\\ {}} & \makecell{ Packing blocks \\ (\textbf{unseen colours})}  &  \makecell{Pushing piles \\ (\textbf{unseen task})} \\
  \end{tabular}
  \caption{\textbf{Real-robot experiments.} 
  We report the \textcolor{plotblue}{\rule{0.2cm}{0.2cm}} \textcolor{plotblue}{perceptual score} and \textcolor{plotgray}{\rule{0.2cm}{0.2cm}} \textcolor{plotgray}{physical score}, which indicate the success rates for affordance perception and physical robotic tasks, respectively. The tasks of \textit{Opening drawer} and \textit{Pushing piles} are not included in the training demonstrations.
  }
  \label{fig:real_exps_10}
  \vspace{-10pt}
\end{figure}

\subsection{Evaluation} 
In \textit{Ravens and OOPP}, the model outputs SE(2) tabletop actions. In Tab.~\ref{tab:multi}, our method outperforms other pre-training-based models and even the method designed for pick-and-place. Our method shows strong capability in interpreting actions with real-scanned objects, as demonstrated in the seen and unseen categories in Tab.~\ref{tab:multi} and all  intra- and inter-class generalisation tasks in Tab.~\ref{tab:oopp}. Our model shows limited performance in coloured geometries, possibly because they are less common in real-world scenarios in our pre-training data. 

In \textit{Franka Kitchen}, as in Tab.~\ref{tab:combined-results} (b), our method achieves state-of-the-art results even compared with supervised pre-training methods~\cite{mpi,chenSUGAR2024}. Unlike Ravens and OOPP, the model in \textit{Franka Kitchen} is required to predict the next timestep's joint angles of the robot arm. We hence employ the encoder and fusion module in our model as a frozen backbone and discard the decoder, consistent with other compared baselines.

\textit{Referring expression grounding} results in Tab.~\ref{tab:combined-results} (a) demonstrate that our method achieves better or comparable results, even compared with supervised methods~\cite{mpi, chenSUGAR2024}. This validates that the goal-image prediction pretext task can learn the association between the target object and the corresponding language instructions, resulting in accurate object localisation.

In \textit{real robot experiments}, we evaluate our model on a real robot across ten distinct tasks, as shown in Fig.~\ref{fig:real_exps_10}. Each task contains 5 different scenarios, and we calculate the overall success rates. The performance is consistent with the results observed in the simulation. Notably, it demonstrated strong generalisation to unseen colours, unseen objects, and even previously unseen tasks. In Fig.~\ref{fig:real_exps_10} we report both perceptual score and physical score, to account for the case where the model correctly outputs the affordance but fails during physical execution due to inaccurate control and object movements, showing the perception-to-physical gap in real-world scenarios (see Sec.~\ref{sec.limitations} for details).

\begin{figure*}[tp]
  \centering
  \footnotesize
  \setlength{\tabcolsep}{0.8pt}
  \newcommand{\sz}{0.1}
  \begin{tabular}{lcccc}
    
    % \raisebox{-22pt}{\rotatebox{90}{\makecell{\small Pack lemon \\ \small into brown box}}} &
    \makecell{ \scriptsize\textsf{ \scriptsize Pick the yellow}\\ \scriptsize\textsf{cube to bowl}}&
    \makecell{\includegraphics[height=\sz\linewidth]{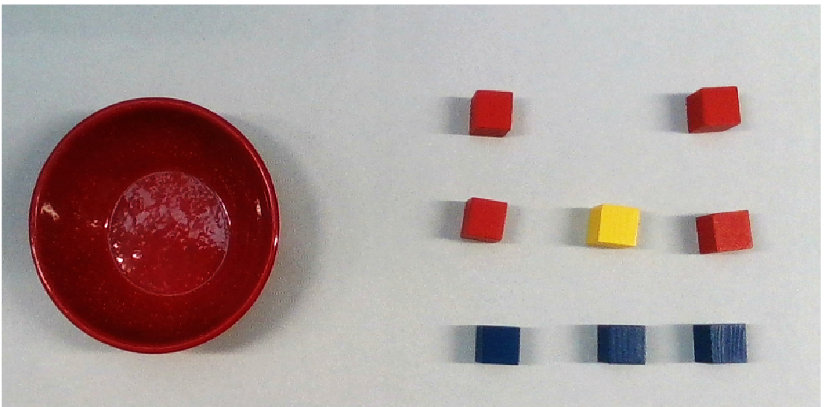}} &
    \makecell{\includegraphics[height=\sz\linewidth]{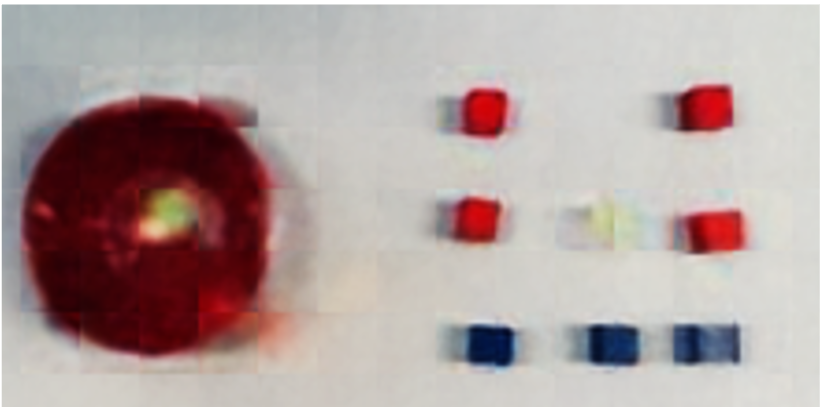}} &
    \makecell{\includegraphics[height=\sz\linewidth]{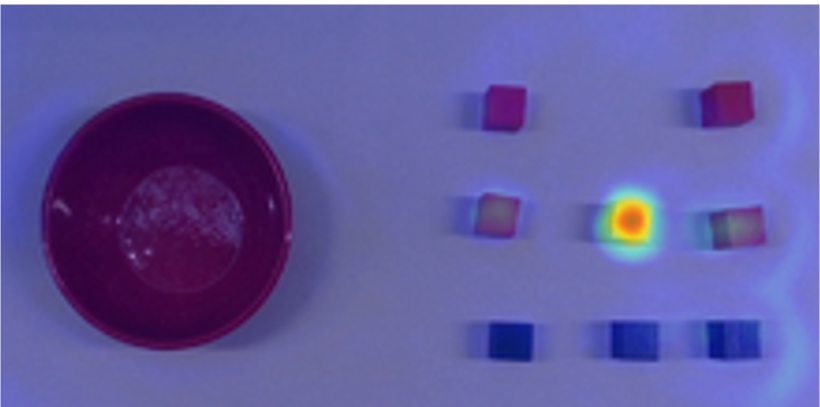}} &
    \makecell{\includegraphics[height=\sz\linewidth]{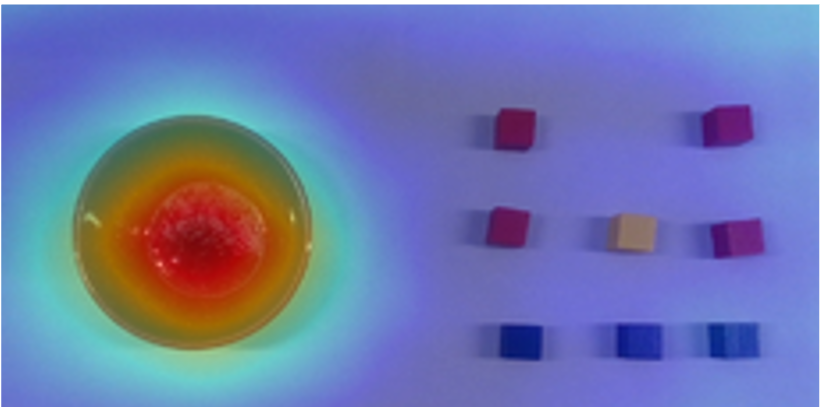}}\\

    % \makecell{Pick the orange \\ball to the bowl}&
    % \makecell{\includegraphics[height=\sz\linewidth]{images/images_grid/3_ori.png}} &
    % \makecell{\includegraphics[height=\sz\linewidth]{images/images_grid/3_predic.png}} &
    % \makecell{\includegraphics[height=\sz\linewidth]{images/images_grid/3_pick.png}} &
    % \makecell{\includegraphics[height=\sz\linewidth]{images/images_grid/3_place.png}}\\
    
    \makecell{\scriptsize\textsf{Fold cloth from} \\ \scriptsize\textsf{left to right}}&
    \makecell{\includegraphics[height=\sz\linewidth]{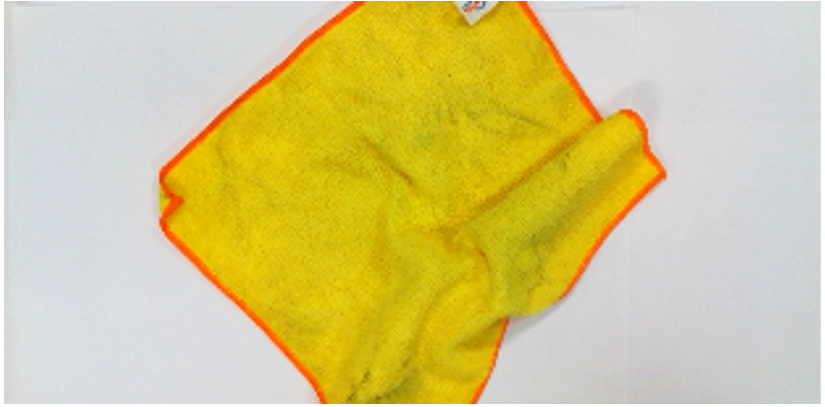}} &
    \makecell{\includegraphics[height=\sz\linewidth]{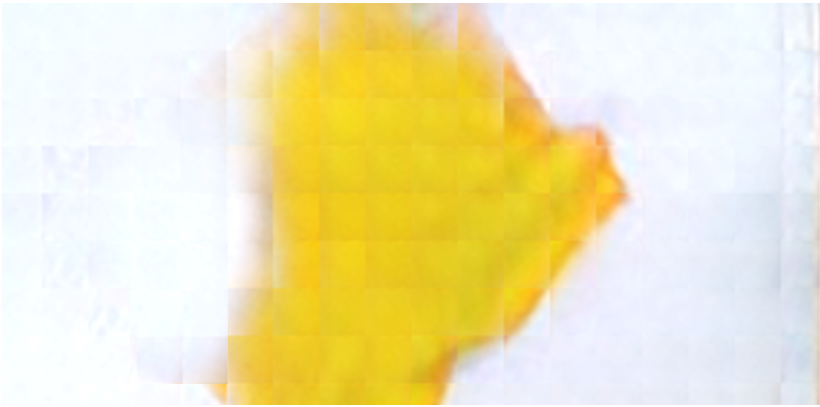}} &
    \makecell{\includegraphics[height=\sz\linewidth]{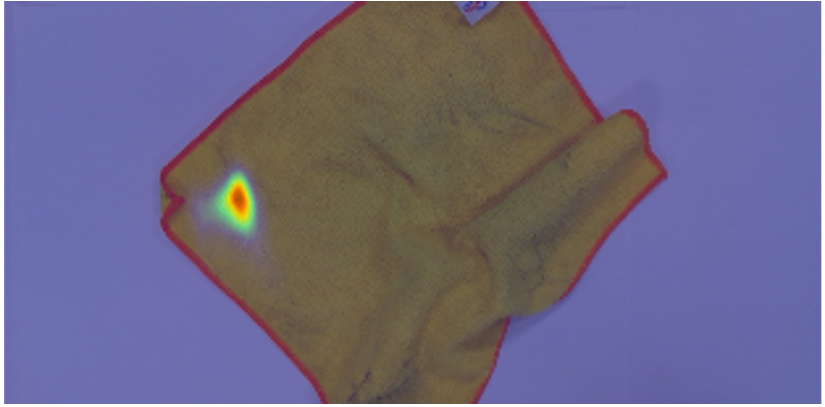}} &
    \makecell{\includegraphics[height=\sz\linewidth]{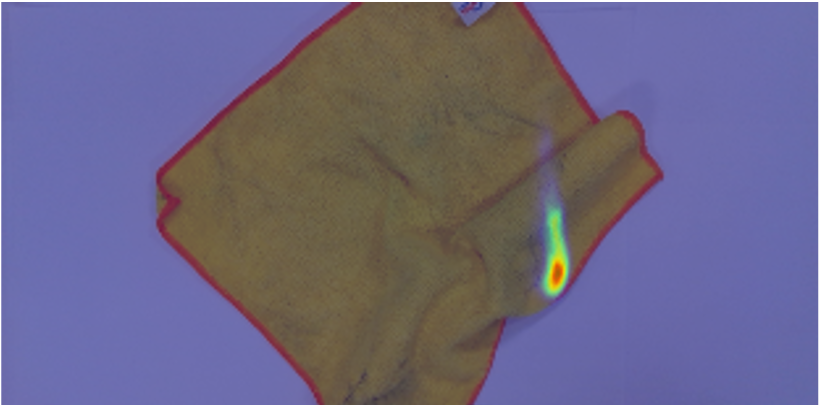}}\\
    % \makecell{Fold cloth from \\bottom to top}&
    % \makecell{\includegraphics[height=\sz\linewidth]{images/images_grid/2_ori.png}} &
    % \makecell{\includegraphics[height=\sz\linewidth]{images/images_grid/2_pred.png}} &
    % \makecell{\includegraphics[height=\sz\linewidth]{images/images_grid/2_pick.png}} &
    % \makecell{\includegraphics[height=\sz\linewidth]{images/images_grid/2_place.png}}
    & Input image & Zero-shot prediction & Pick affordance & Place affordance \\
  \end{tabular} 
  \caption{\textbf{Qualitative examples from real-robot manipulation.} We show qualitative examples in our real-robot environment. Our goal-image predictions are blurry as in other MAE-based methods~\cite{he2022vlmae,weinzaepfel2022croco,guptaSiamese2023}, but still show plausible object placement and the results are aligned with the input text instructions, which leads to accurate manipulation. (Only affordances for translation are shown)}
  \vspace{-8pt}
  \label{fig:qual_comparison}
\end{figure*}

\subsection{Analysis}
\definecolor{color2}{HTML}{8ecae6}
\begin{figure}[t]
\centering
\begin{tabular}{@{}cc@{}}
    \raisebox{40pt}{
    \resizebox{0.4\textwidth}{!}{%
        \begin{tabular}{@{}lccc@{}}
            \toprule
            Method & \makecell{Rotation  \\ tasks}  &  \makecell{Translation  \\ tasks} & Avg. \\ \midrule
            w/o fusion module                 & 82 & 66 & 72 \\
            w/o OOPP data & 87 & 65 & 74 \\
            w/o Asym. mask                      & {\ul 91} & {\ul 68} & {\ul 77} \\
            % use BERT                          & 87 & 61 & 71 \\
            \rowcolor[HTML]{ECF4FF} 
            Ours                              & \bf 96 & \bf 71 & \bf 81 \\ \bottomrule
            \end{tabular}%   
        }%
    }
    &
    \resizebox{0.5\textwidth}{!}{
    \begin{tikzpicture}
    \begin{axis}[
        width=0.5\linewidth,
        height=3.5cm, 
        ylabel={\footnotesize Success rate ↑},
        xmode=log,
        log basis x={10},
        log ticks with fixed point,
        xtick={0.75,0.85,0.90,0.95,1.00},
        ymin=0.5, ymax=1.0,
        grid=both,
        legend style={font=\scriptsize, at={(0.95,0.05)}, anchor=south east},
        tick label style={font=\fontsize{7}{8}\selectfont},
        label style={font=\footnotesize},
        xlabel style={at={(axis description cs:0.5,-0.1)}},
        ylabel style={at={(axis description cs:-0.08,0.5)}},
        line width=1pt,
        mark size=2pt,
    ]
    
    % CLIPort line
    \addplot+[dashed, mark=*, color=color2, mark options={solid, draw=color2, fill=color2}] coordinates {
        (0.75, 0.66)
        (0.85, 0.69)
        (0.90, 0.79)
        (0.95, 0.83)
        (1.00, 0.63)
    };
    %\addlegendentry{CLIPort}
    
    \end{axis}
    \end{tikzpicture}} \\
    
    {\footnotesize (a) Effect of pre-training strategy} & {\footnotesize (b) Effect of the masking ratio}
\end{tabular}
\caption{\textbf{Ablation analysis.} We analyse the effect of pre-training strategies and masking ratios on the Franka Kitchen benchmark and Ravens benchmark,  respectively.}
\label{fig:ablation}
\end{figure}
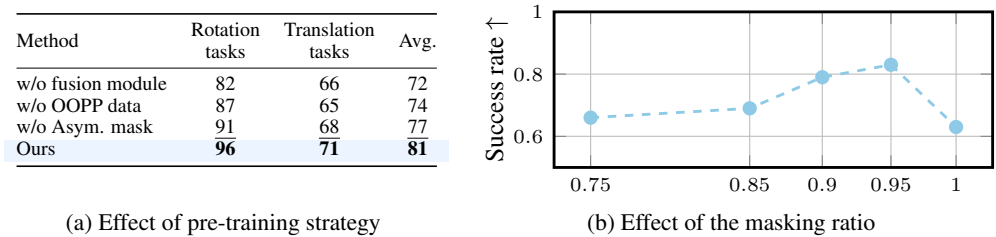

In Fig.~\ref{fig:ablation} (a), we present a set of ablation studies on the Franka Kitchen benchmark to evaluate the contribution of several key components of our approach.
We evaluate the effect of removing the fusion module, which injects the language embedding directly into the decoder via cross-attention. The suboptimal results are consistent with other recent works~\citep{liuGrounding2023,glip}, showing the importance of integrating visual and textual modalities before decoding.
We also examine the impact of training with the proposed OOPP dataset, which introduces a wide variety of object classes. Results show that the downstream performance can be significantly boosted, indicating the model benefits from broader object priors in the OOPP dataset to learn generalisable visual-action representations.

Finally, we evaluate the effect of the asymmetric masking strategy and the optimal masking ratio. Without asymmetric masking, the model may not be able to fully capture the underlying causality in language-guided manipulation, leading to reduced accuracy in downstream tasks, as shown in Fig.~\ref{fig:ablation} (a). In Fig.~\ref{fig:ablation} (b), we further study the impact of different masking ratios by training models under four commonly used settings~\cite{heMasked2022, weinzaepfel2022croco, tongVideoMAE2022}. A 95\% masking ratio performs the best. Lower ratios leak too much information, making learning less effective, while a 100\% ratio, where prediction relies entirely on input, leads to convergence issues due to the future state's ambiguity. 

\section{Conclusion}

We presented a visual-action representation learning framework for robot manipulation that leverages goal-image prediction as a pretext task to capture the underlying causality in language-guided robot manipulation. To facilitate learning from diverse object priors, we also proposed the OOPP dataset, which provides a rich collection of tabletop manipulation episodes. Through extensive evaluations in both simulated and real-world settings, our method demonstrates superior performance over existing baselines, exhibiting strong generalisation and effective knowledge transfer across tasks. Future work includes scaling training to diverse video datasets to learn a robust and generalisable representation and extend our learned representations for more complex robotic tasks.

%==========================================================================

%==========================================================================

% \section{Citations}
% \label{sec:citations}

% 	Citations can be made using either \textbackslash citep\{\} or \textbackslash citet\{\}, depending from the appropriateness. To avoid the citation moving to the next line, it is often a good practice to replace the space before with a tilde (\~{}) character.
% 	Example 1: ``CoRL is the best conference ever~\citep{Gauss1857}.''
% 	Example 2: ``\citet{Lagrange1788} proved, both theoretically and numerically, that CoRL is the best conference ever.''
	
%===============================================================================
% =================================================================

\newpage
\section{Limitations}
\label{sec.limitations}
\begin{figure*}[h]
  \centering
  \footnotesize
  \setlength{\tabcolsep}{1.pt}
  \newcommand{\sz}{0.12}
  \begin{tabular}{lccc}
    \makecell{\textsf{\scriptsize Pick the yellow}\\ \textsf{\scriptsize ring to the middle}}&
    \makecell{\includegraphics[height=\sz\linewidth]{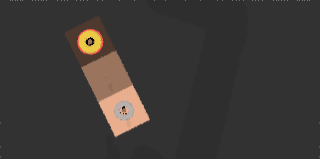}} &
    \makecell{\includegraphics[height=\sz\linewidth]{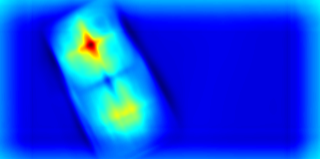}} &
    \makecell{\includegraphics[height=\sz\linewidth]{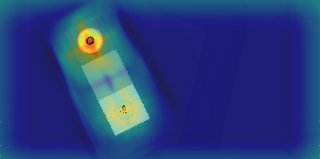}}\\
  \end{tabular} 
  \caption{Visualised affordance of \textit{Towers-of-Hanoi} task from the top-down view in Ravens.}
  \label{fig:fail_affordance}
\end{figure*}

\boldparagraph{Manipulating precision.} 
Our model exhibits limited performance when dealing with small objects or those with fine-grained structures, such as in \textit{Towers-of-Hanoi} and \textit{Stack-Block-Pyramid} (Tab.~\ref{tab:multi}). This is attributed to the use of ViTs, which process images as patches and may overlook fine details that are important for accurate manipulation, as shown in Fig~\ref{fig:fail_affordance}. Recent methods that enhance the spatial resolution by learning to upsample ViT features, such as DPT~\cite{dpt} or FeatUp~\cite{featup},  could potentially mitigate this issue and are promising directions for future exploration.

\boldparagraph{Pseudo affordance.} 
On the benchmark of Ravens~\cite{zeng2021transporter} and OOPP, our method predicts a pseudo-affordance map where the pixel with the highest value is treated as the manipulation point (See Fig.~\ref{fig:fail_affordance}). While this approach simplifies fine-tuning, it can be problematic for objects with complex geometry. In such cases, the manipulation can fail due to subtle physical interactions even if the predicted pick-and-place poses are geometrically correct. This leads to a discrepancy between the perceptual score and the actual physical success rate, as illustrated in Fig.~\ref{fig:real_exps_10}.  Incorporating self-supervised learning that accounts for human or robot interaction dynamics could help mitigate this issue by enabling the model to learn priors over common manipulable regions.

\boldparagraph{3D awareness.} Our model lacks explicit 3D understanding and relies solely on 2D visual cues. As a result, it can struggle to distinguish between objects with similar textures or appearances but differing 3D shapes or structures. Incorporating self-supervised learning methods that promote geometric awareness may help the model learn 3D priors and improve robustness in such cases.

\boldparagraph{Articulated objects.}
In real-world scenarios, some objects are composed of multiple rigid parts and interactable. However, our visual pre-training primarily focuses on rigid object manipulation and does not explicitly model object-part interactions. Including articulated objects with several new tasks in our OOPP dataset can potentially help capture these complex interactions.

\boldparagraph{Blurry prediction results.} MAE-based methods are known to produce blurry reconstructions lacking high-frequency details~\cite{guptaSiamese2023, weinzaepfel2022croco, he2022vlmae}. Our goal image prediction also suffers from the blurry issue, as shown in Fig~\ref{fig:teaser}. However, our objective is not to generate high-quality images, but rather to use goal image prediction as a pretext task for learning semantic representations. In addition, blurriness reflects the missing of high-frequency components in the image, which often correspond to perceptual details rather than semantic contents. In fact, despite the perceptual blur, the preserved semantic structure in our predictions demonstrates that the model captures semantic information effectively. The lack of perceptual fidelity may limit interpretability for humans, and we consider this as future work.

\section{Acknowledgement}
This work was supported by the Royal Society Research Grant RGS/R2/242051, and utilised the Sulis Tier 2 HPC under the UK EPSRC Grant EP/T022108/1 and the HPC Midlands+ consortium.

\clearpage
\bibliography{main}  % .bib

\clearpage
\begin{center}
    \LARGE \textbf{Appendix}
\end{center}

\pagenumbering{arabic} 
\setcounter{page}{1}
\setcounter{section}{0}

\section{Introduction}
We provide additional material that supports our paper.

\begin{itemize}[itemsep=3pt, parsep=0pt, topsep=0pt, partopsep=0pt, leftmargin=*]
\item We invite readers to visit our website about the proposed pretext task (goal-image prediction), Omni-Object Pick-and-Place (OOPP) dataset, and real and simulated robot manipulation.
\item In Sec.~\ref{sup:addtional_details}, we provide more details about 
our method, including our backbone and different output heads utilised in different downstream tasks.
\item In Sec.~\ref{sup:benchmark}, we describe more details of each downstream task, including the real-robot experiment.
\item In Sec.~\ref{sup:additional_results}, we provide additional examples of the proposed pretext task, goal-image prediction, and our OOPP dataset.

\end{itemize}

\section{Method details}
\label{sup:addtional_details}

\boldparagraph{Backbone architecture.}
\label{sup:network_details}
We adopt ViT-Base~\cite{dosovitskiy2020image} with a patch size of 16 as our backbone. The original Vit-Base architecture consists of 12 attention blocks with an embedding dimension of 768, and its the decoder typically comprises of 8 attention blocks with an embedding dimension of 512. For a fair comparison with other methods using ViT-Base as a backbone, we include 6 attention blocks and 6 bi-directional attention blocks in our encoder. In the pre-training stage, our goal-image prediction head is a one-layer fully connected network.

\boldparagraph{Affordance head.}
Our affordance head for generating $\mathbf{SE}(2)$ robot actions (the 2D location and 1D rotation) for tabletop manipulation tasks (\textit{Ravens}, \textit{OOPP}, and our real robot experiments) consists of 4 convolutional layers with skip connection to convert the output from the transformer backbone to the affordance map. To generate a rotation angle, we expand a single affordance map into 36 instances, where each instance represents a 10-degree step. We apply the softmax function to the expanded affordance map, which identifies the position and corresponding rotation angle. Since other pre-training methods have not been previously evaluated on this benchmark, we reimplemented them and used a unified action head across all pre-training baselines. For other Pick-and-Place methods, we retain their original action heads which are structurally similar but typically deeper due to their use of ResNet backbones. In addition, since our pre-training methods are designed to predict the goal image, we found that concatenating the goal image into the convolutional layers facilitates improved affordance generation, as the two tasks are closely correlated. 

\boldparagraph{Action head and detection head.}
\label{sec:action_head}
Following the baseline established by Voltron~\cite{karamchetiLanguageDriven2023} and MPI~\cite{mpi}, we adopt the same shallow MLP policy network to predict joint velocities of \(\mathbb{R}^9\) (7 degree of freedom and two for grasp status) for robot actions in the Franka Kitchen benchmark, and to regress bounding boxes in the Referring Expression Grounding benchmark. We use features after the fusion module for two main reasons: First, for fair comparison: previous methods, including R3M~\cite{r3m}, MVP~\cite{mvp1} and Voltron~\cite{karamchetiLanguageDriven2023}, only evaluate frozen representations dropping the decoder part. Since we directly report their results on these benchmarks, we adopt a consistent setting. Second, based on task requirements, both benchmarks rely on understanding the current state or the next state, while features after the decoder in our model represent the goal image and correspond to the final state. Therefore, using features before the decoder is more suitable for these two benchmarks.

\section{Benchmark details}
\label{sup:benchmark}
\subsection{Ravens}

\begin{figure*}[b]
  \centering
  \includegraphics[width=0.95\linewidth]{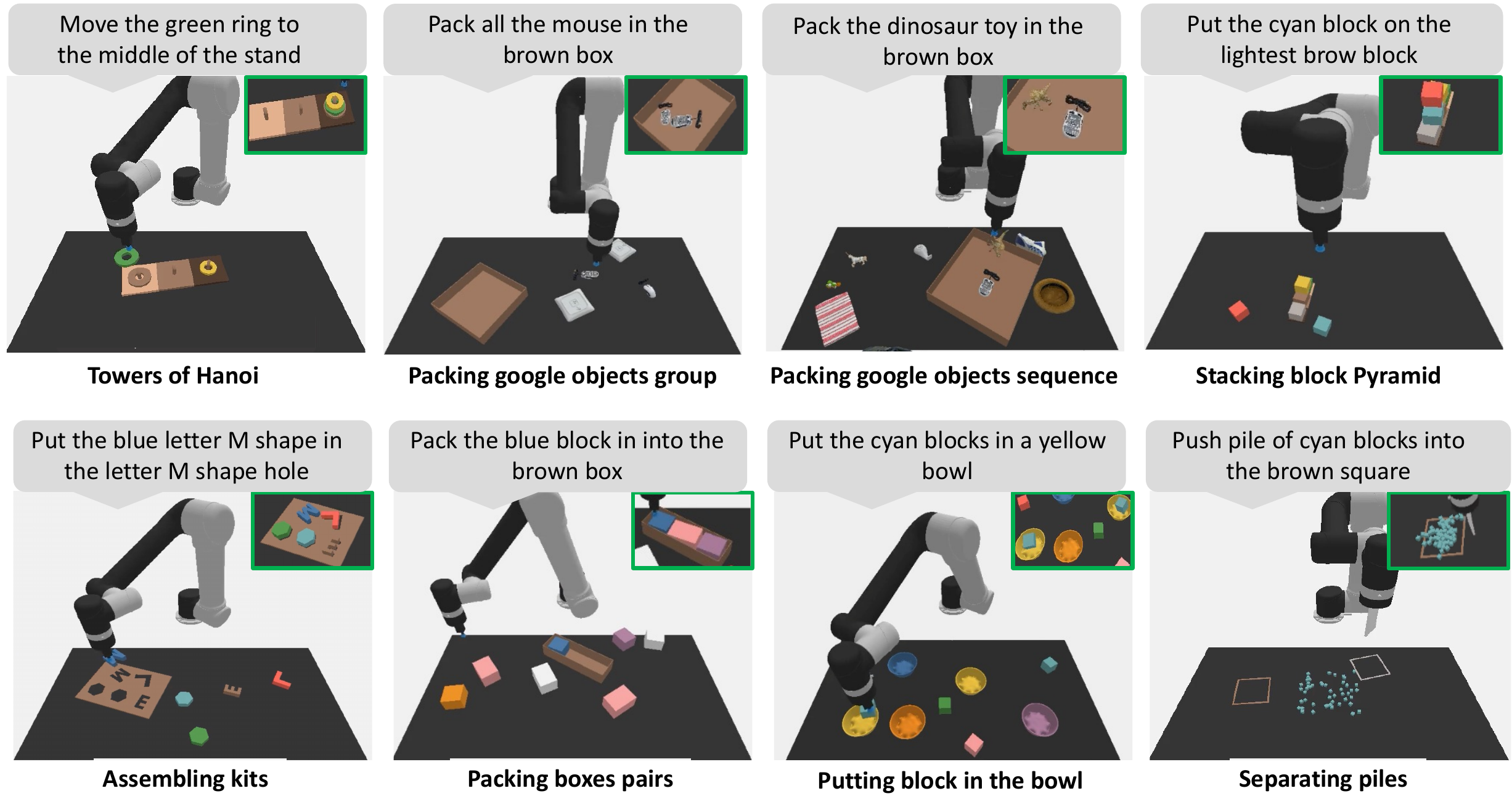}
  % \vspace{-5pt}
  \caption{\textbf{Examples of eight robot manipulation tasks in the simulator.} The language instructions are on the top of each image and the final states are shown in the green box.}
  \label{fig:exp_sim_task}
  % \vspace{-5pt}
\end{figure*}

\boldparagraph{Overview.} Ravens~\cite{zeng2021transporter} is a simulated benchmark to evaluate tabletop pick-and-place robot manipulation tasks.  We use PyBullet OpenAI Gym~\cite{benelot2018} based on the configuration described in CLIPort~\cite{shridhar2022cliport}. We chose 8 language-conditioned tasks for the experiment, as shown in Fig.~\ref{fig:exp_sim_task}, including \textit{Packing seen or unseen Google objects sequence}, \textit{Packing seen or unseen Google objects group}, \textit{Put block in the bowl}, \textit{Stack blocking pyramid}, \textit{Towers of Hanoi}, \textit{Packing boxes pairs}, \textit{Assembling kits}, and \textit{Separating piles}. For details of each task, including the train and test split of objects, and the language instruction template, please refer to~\cite{shridhar2022cliport, zeng2021transporter}. Note that we did not split the tasks according to seen or unseen colours, as all the methods have already ``seen" all the colours in their pre-train models or the unsupervised pre-training phase. Therefore, we combine both ``seen" and ``unseen" splits of colours into a single task and the scores reflect the model's perception ability on all the colours.

\boldparagraph{Evaluation details.} We evaluated the capability of the proposed methods on multi-task experiments in the benchmark. Specifically, we trained the model using 1,000 demonstrations drawn from all task categories and assessed performance on another 100 test demonstrations per task. Since prior pre-training methods~\cite{karamchetiLanguageDriven2023, mpi} have not been evaluated on this benchmark, we reimplemented their models using the original codebases and pre-trained them on the same pre-training data as our approach. When adapting to the downstream tasks, these pre-training methods were also equipped with the affordance head described in Sec.~\ref{sec:action_head}. Subsequently, all baselines, including the Pick-and-Place baselines, were fully fine-tuned on the same set of downstream demonstrations, following the evaluation protocol outlined in~\cite{shridhar2022cliport}, to ensure a fair comparison.

\subsection{Franka Kitchen}
\begin{figure*}[tb]
  \centering
  \scriptsize
  \setlength{\tabcolsep}{0.8pt}
  \newcommand{\sz}{0.185}
  \begin{tabular}{ccccc}
    \makecell{\includegraphics[height=\sz\linewidth]{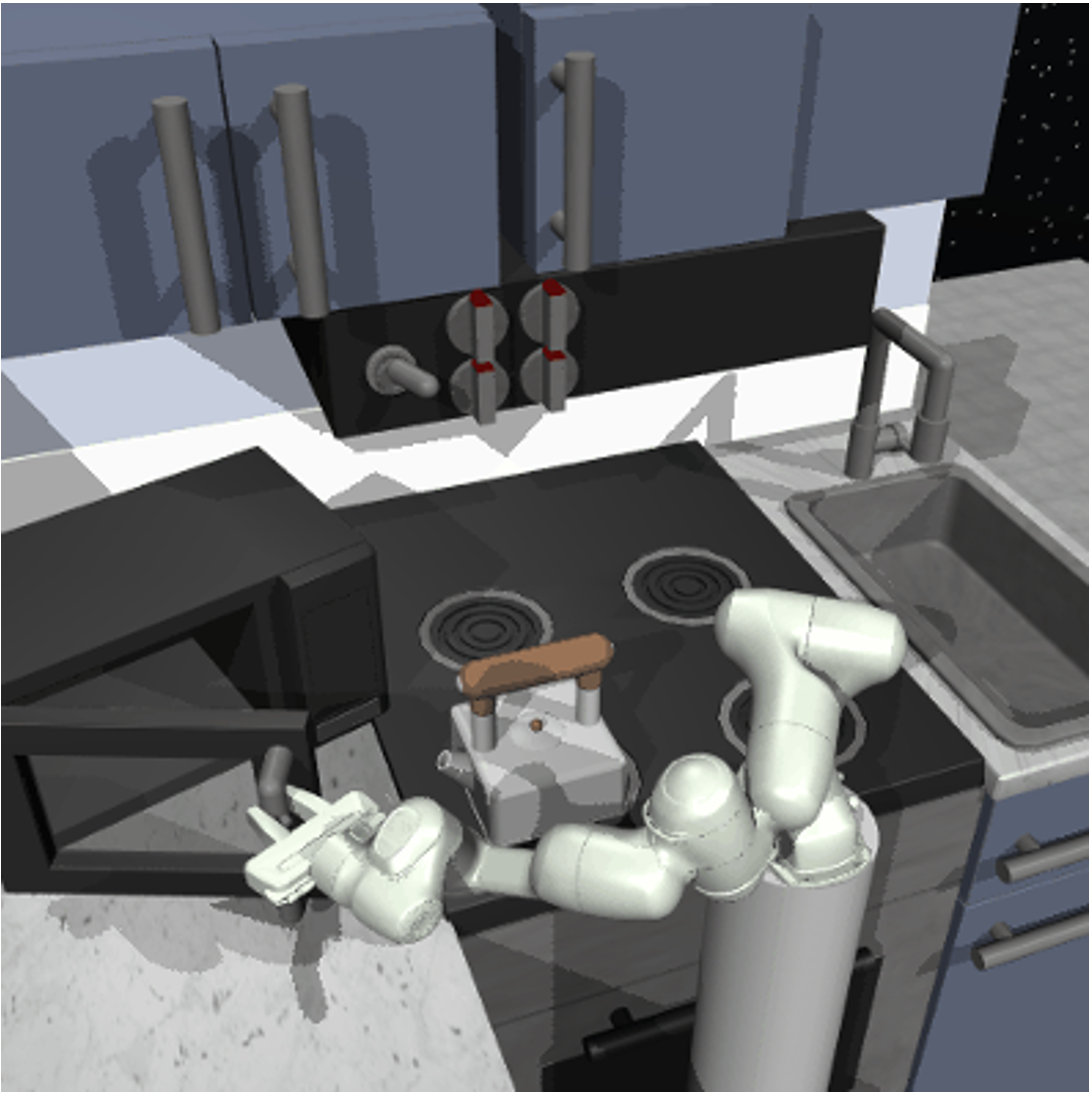}} &
    \makecell{\includegraphics[height=\sz\linewidth]{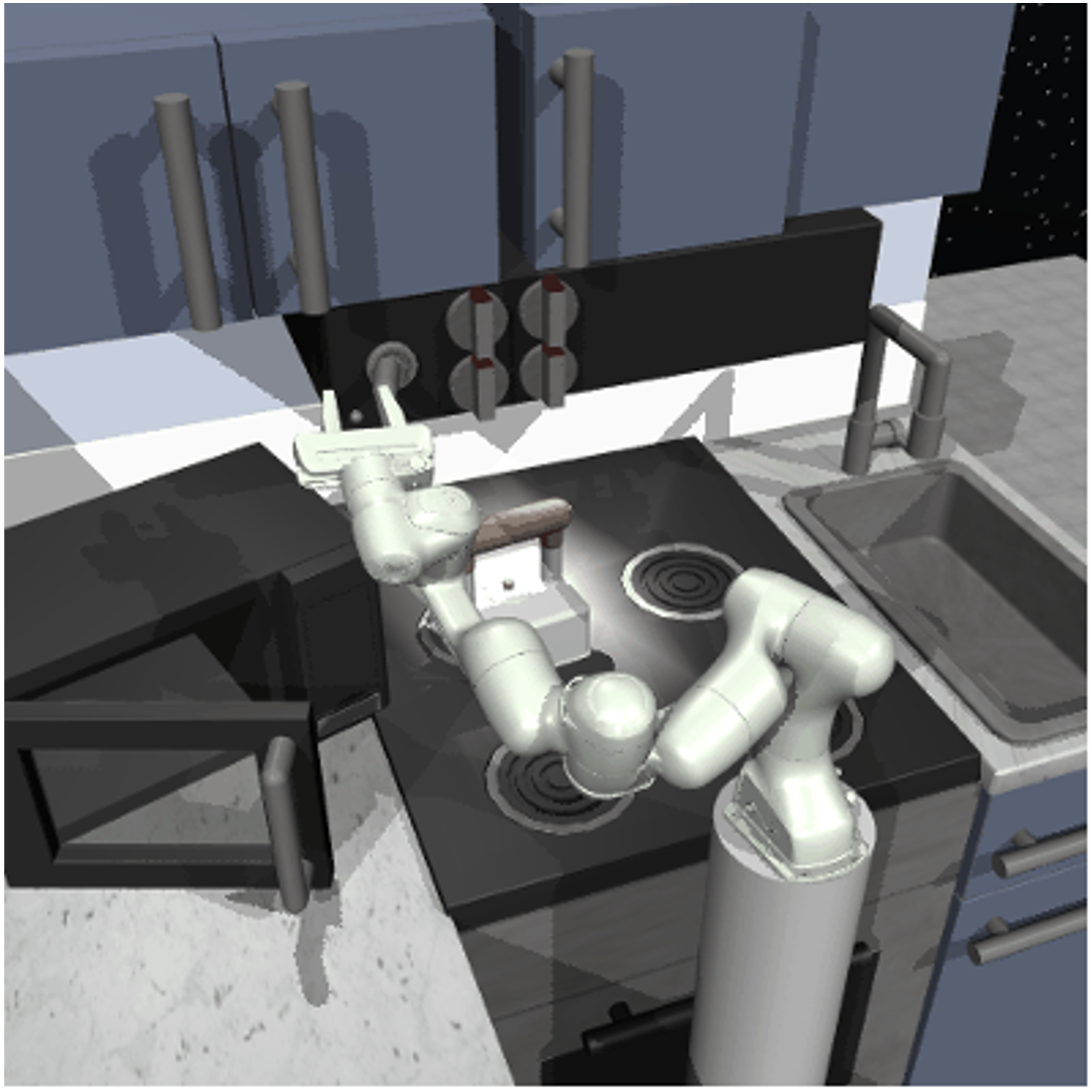}} &
    \makecell{\includegraphics[height=\sz\linewidth]{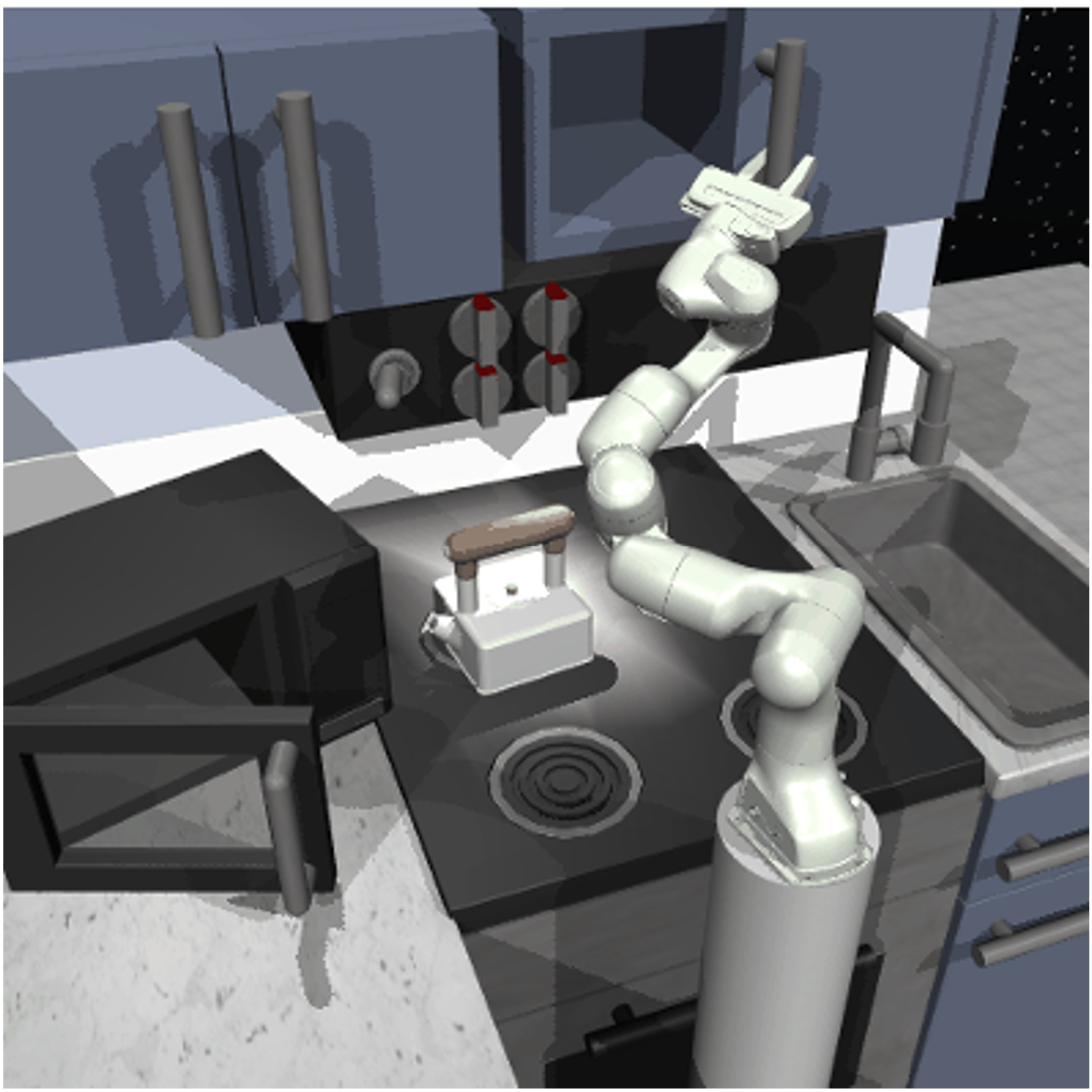}} &    
    \makecell{\includegraphics[height=\sz\linewidth]{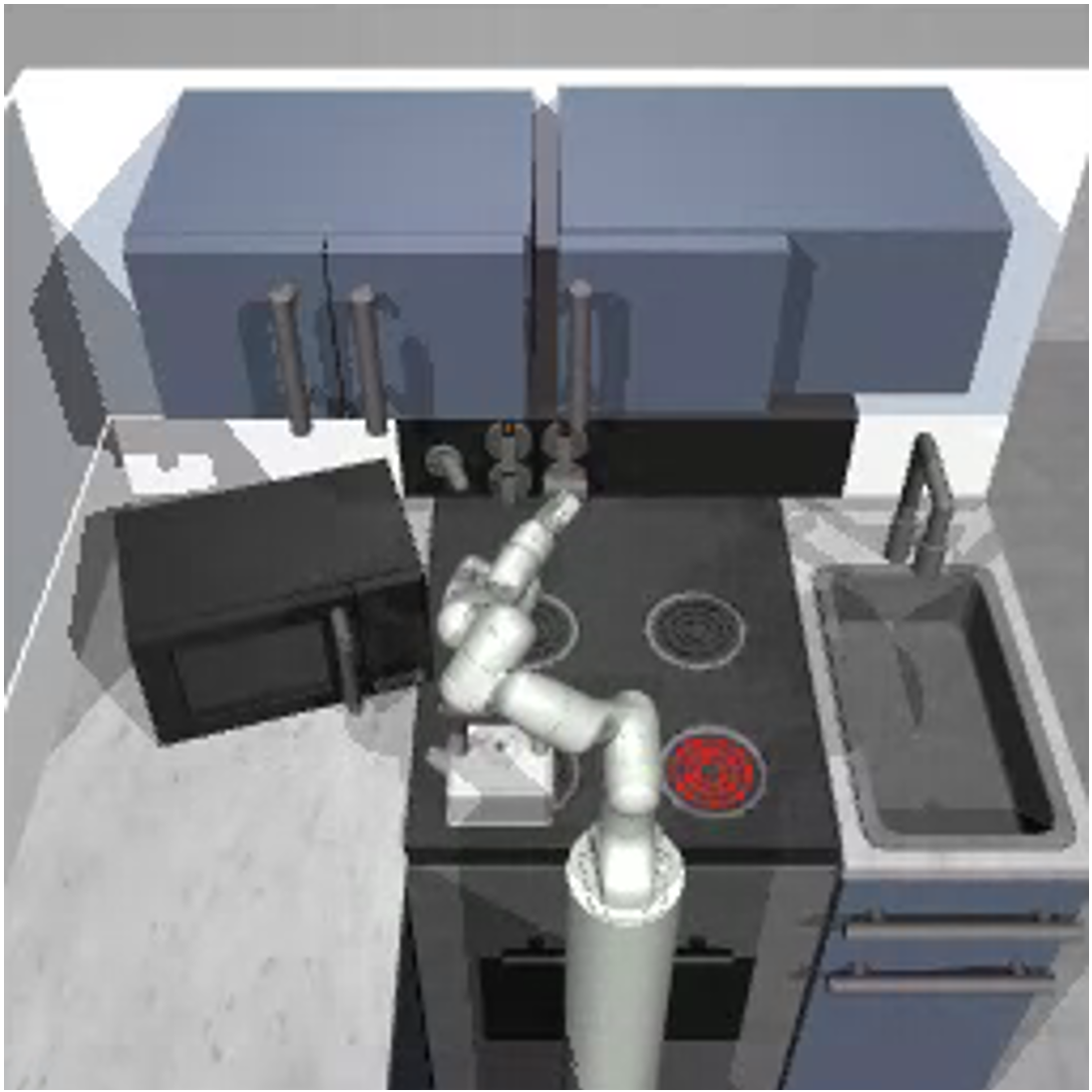}} &
    \makecell{\includegraphics[height=\sz\linewidth]{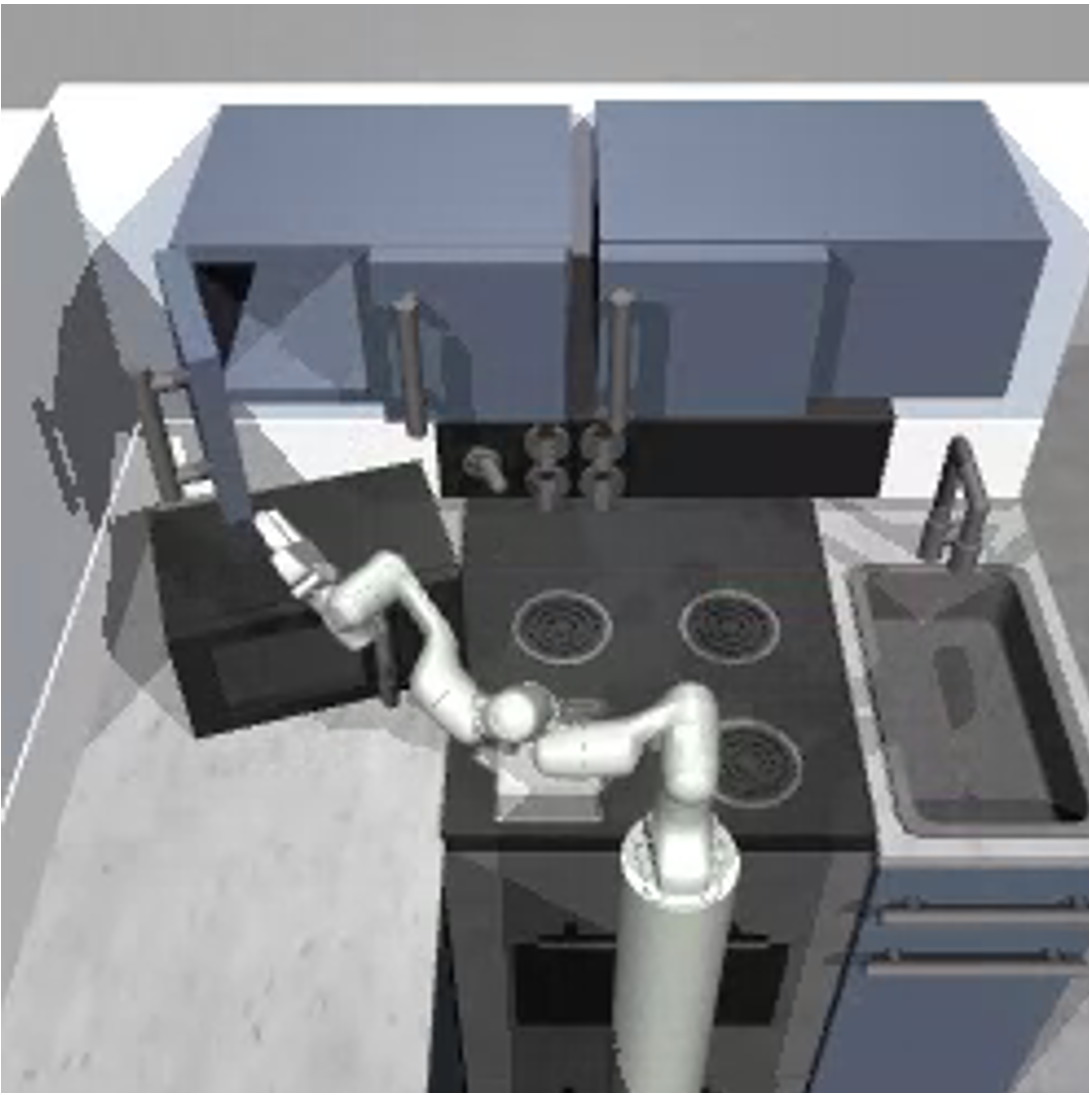}} \\
    Open the microwave & Turn on the light & Slide the right door & Turn the stove top knob & Open the left door\\
  \end{tabular} 
  \caption{\textbf{Tasks in the Franka Kitchen benchmark.}}
  \label{fig:franka_examples}
\end{figure*}

\boldparagraph{Overview.} Franka Kitchen~\citep{franka_kitchen} is a well-established benchmark for evaluating the efficacy of visual representations in facilitating the learning of visuomotor control policies from limited demonstrations. This benchmark comprises five distinct visuomotor control tasks, as shown in Fig.~\ref{fig:franka_examples}, each captured from two camera viewpoints. 

\boldparagraph{Evaluation details.} We use the action head described in Sec.~\ref{sec:action_head} for predicting joint velocities. As prior works~\citep{radfordLearning2021a, karamchetiLanguageDriven2023, mvp1, mpi, r3m}, which leverage supervised or unsupervised pre-training for robot manipulation, commonly adopt this benchmark, we directly report the results stated in their original papers and compare our method against these approaches. Following the evaluation protocols widely adopted in these works, we trained the action head with the backbone frozen using 25 demonstrations, and report average success rates across five tasks, two viewpoints, and three random seeds.

\subsection{Referring Expression Grounding}
\boldparagraph{Overview.} The goal of this task is to predict a bounding box of an object in a cluttered scene based on the nature language expression. This task offers the evaluation of language-conditioned scene understanding and object recognising ability, which serves as an important prerequisite for language-based robot manipulation. The benchmark is based on OCID-Ref Dataset~\cite{ocid}, which provides representatives scenes in robotics settings. The benchmark also provides splits based on the clutter level. 

\begin{figure}[b]
  \centering
  \includegraphics[width=0.9\linewidth]{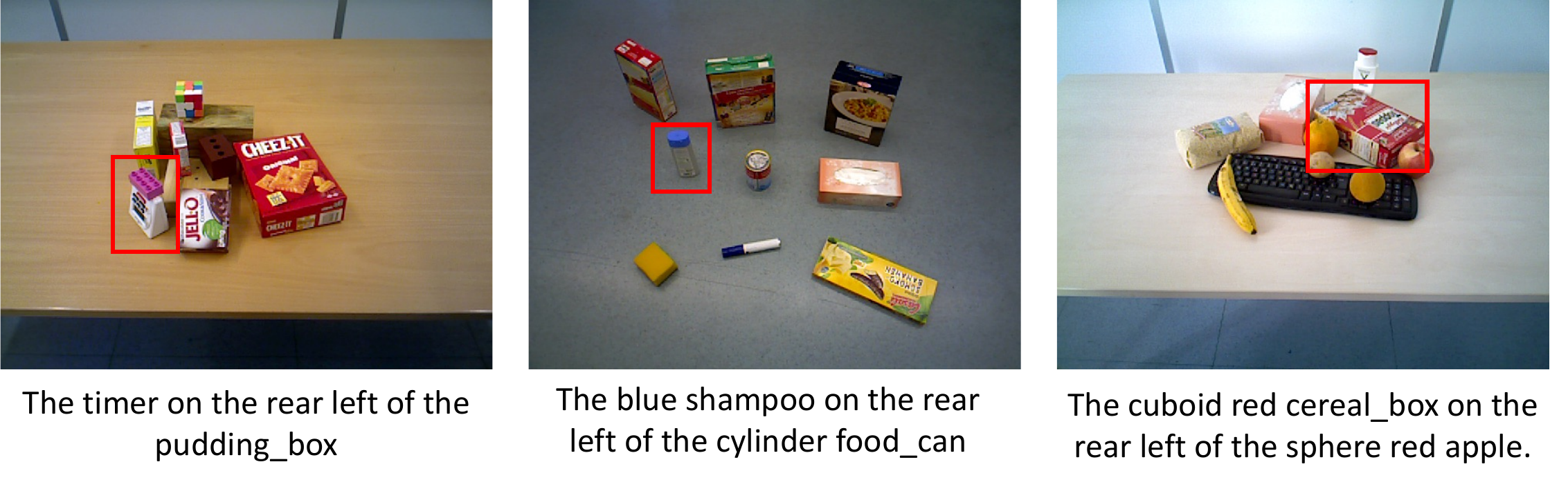}
 \caption{\textbf{Examples of the task of Referring Expression Grounding.} This task offers varied language instructions involving a wide range of objects and scenes.}
  \label{fig:heads}
\end{figure}

\boldparagraph{Evaluation details.} We use a shallow MLP as detection head to regress the bounding box directly. We use the evaluation codebase provided by~\cite{karamchetiLanguageDriven2023}. Similar to Franka Kitchen, we report the results stated in the paper for each baseline~\cite{karamchetiLanguageDriven2023, mvp1, mpi, chenSUGAR2024, r3m}. The evaluation metrics are the average precision at 0.25 IoU under each clutter level.

\subsection{Real robot experiments}
\label{sup:real-word_setup}
\boldparagraph{Overview.}
We validate the applicability of our method in real-world scenarios. We validate our model on 10 manipulation tasks: \textit{Stacking blocks}, \textit{Folding cloth}, \textit{Packing objects}, \textit{Opening drawer}, \textit{Pressing button}, \textit{Aligning rope}, \textit{Packing blocks}, and \textit{Pushing piles}. Each task contains 5 different scenarios that differ in either objects or locations. We manually collected 200 training demos that contain robot image pairs, language descriptions, and annotations for real-world fine-tuning. Five colored blocks and five unseen objects were excluded from the training demonstrations, and no demonstrations from the \textit{Opening drawer} or \textit{Pushing piles} tasks were included, although similar tasks are present in the images used in the self-supervised pre-training phase. This design aims to evaluate whether the model can effectively generalised from the self-supervised pre-training, rather than relying on downstream demonstrations.

\begin{figure*}[tb]
  \centering
  \setlength{\tabcolsep}{1pt}
  \begin{tabular}{ccc}
    \makecell{\includegraphics[height=0.28\linewidth]{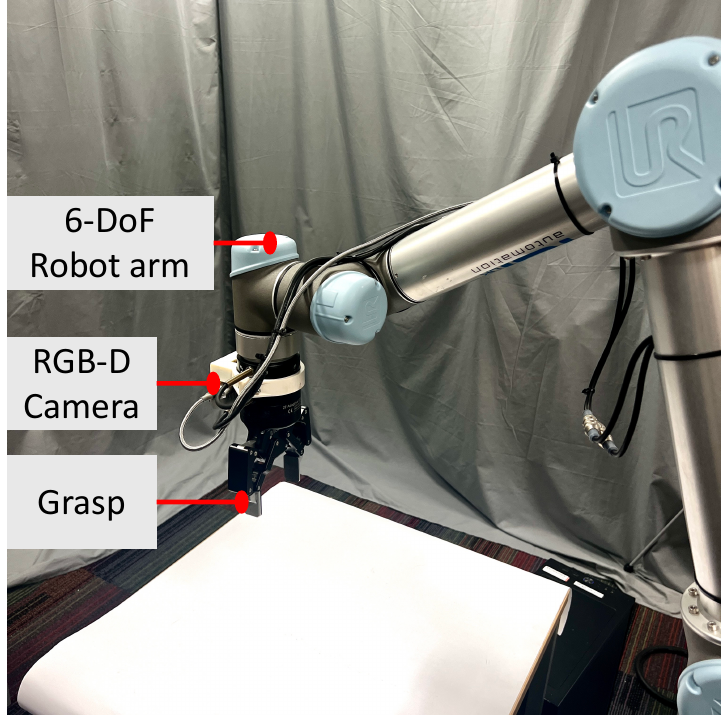}} &
    \makecell{\includegraphics[height=0.28\linewidth]{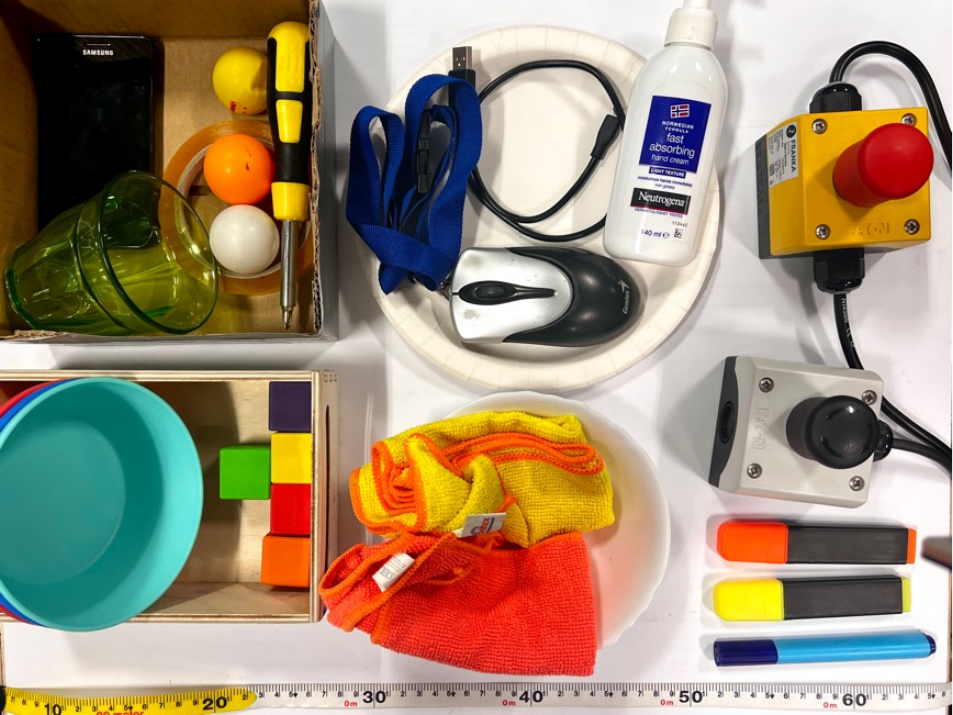}} &
    \makecell{\includegraphics[height=0.28\linewidth]{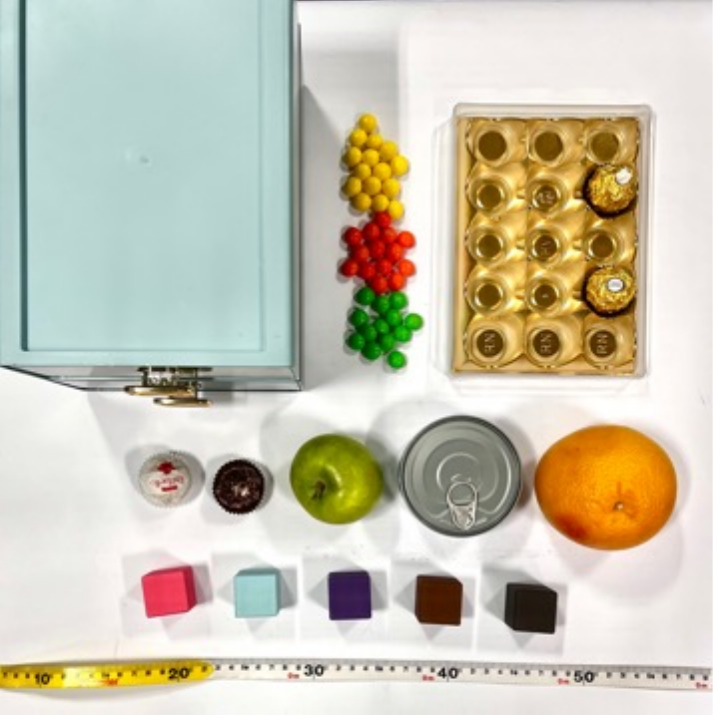}} \\
    Real robot setup & Seen objects and distractors & Unseen objects\\
  \end{tabular} 
  \caption{\textbf{Real robot experiment.} The figure shows our physical robot setup along with both seen and unseen objects. Seen and unseen colored blocks are also included.}
  \label{fig:real_robot_objs}
\end{figure*}

\boldparagraph{Evaluation details.} Fig.~\ref{fig:real_robot_objs} shows our real-robot environment and objects. We train both our model and CLIPort~\cite{shridhar2022cliport} on our manually collected training demos.  We utilise a 6-degree-of-freedom (6-DoF) UR5 robotic arm, Robotiq 2F-85 two-finger gripper, and an Intel RealSense RGB-D camera for our real-world experiments. We capture the top-down RGB observation which covers the workspace of 60 cm $\times$ 30 cm, and the image from the camera is resized to 320 $\times$ 160 pixels. 

\boldparagraph{Task details.} Here we show the language template, variable factors and the success condition for each real robot task in Tab.~\ref{tab:tasks_description}. Images for each task are presented in the main paper.

\begin{table}[]
\centering
\caption{\textbf{Real robot tasks settings.} We present the language template, variable factors and the success condition for each real robot task}
\label{tab:tasks_description}
\resizebox{\columnwidth}{!}{%
\begin{tabular}{llll}
\toprule
\textbf{Task name} & \textbf{Language template} & \textbf{Variable factors} & \textbf{Success condition} \\
\midrule
Stacking blocks &
  Stack the \{color\} block on the \{color\} blocks &
  Block color and position (pick/place) &
  Correct block stacked on target block \\
  
Folding cloth &
  Fold the cloth from \{direction\} to the \{direction\} &
  Cloth color, initial orientation &
  Grasp and fold directions match the template \\
  
Packing objects &
  Pick the \{object\} into the \{object\} &
  Object type, distractor objects &
  Target object placed into correct container \\
  
Opening drawer &
  Pull out the drawer &
  Drawer position&
  Drawer fully opened \\
  
Pressing button &
  Press the \{color\} button &
  Button color, button location &
  Correct button touched \\
  
Aligning rope &
  Align the rope from \{direction\} to the \{direction\} &
  Rope position, direction variation &
  Rope aligned from start to target direction \\
  
Packing blocks &
  Put the \{color\} block into the \{color\} bowl &
  Block color, bowl color &
  Correct block placed into matching bowl \\
  
Pushing piles &
  Push the pile of \{objects\} to the \{color\} area &
  Object color, target area color &
  Pile reaches target area within five attempts \\
\bottomrule
\end{tabular}%
}
\end{table}

\begin{figure*}[t]
  \centering
  \includegraphics[width=0.98\linewidth]{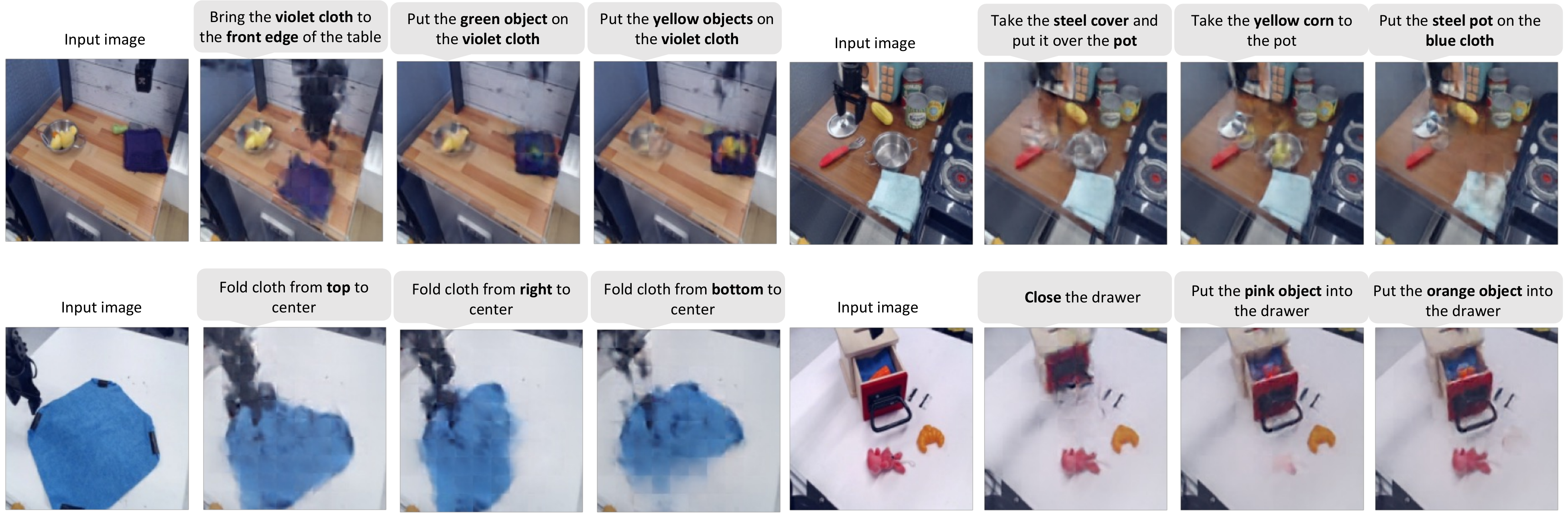}
  \caption{\textbf{Examples of goal prediction images.} Given the same input image with a different language, the model effectively understands the diverse language instructions and predicts goal images aligning with the semantics. Notably, all images belong to the test set and are novel to the model.}
  \label{fig:predictions}
\end{figure*}

\section{Additional details}
\label{sup:additional_results}

\begin{figure}[t]
  \centering
  \includegraphics[width=1.0\linewidth]{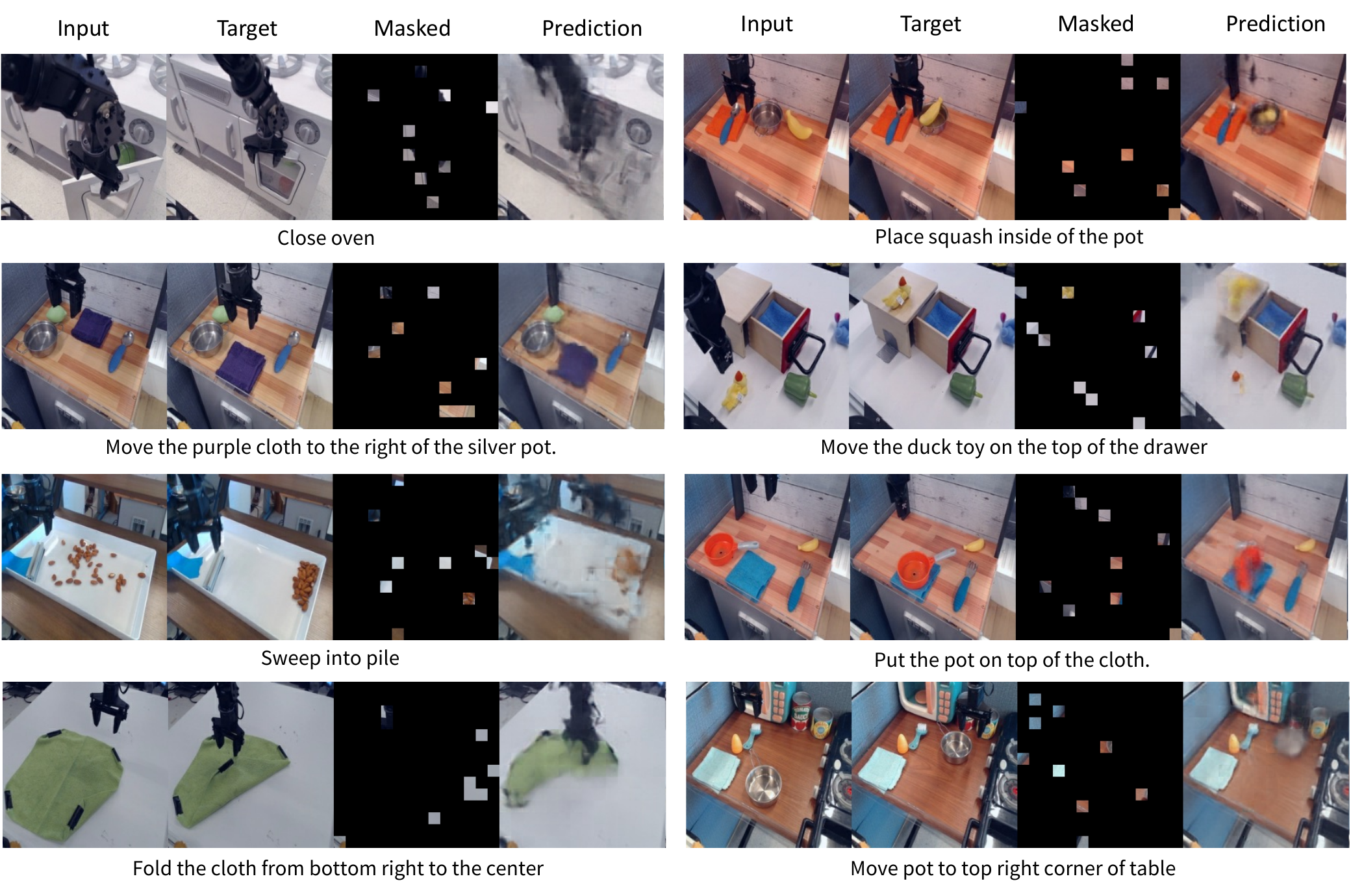}
 \caption{\textbf{Examples of masked image prediction.} We show the qualitative examples of our masked auto-encoders on the validation set. The text descriptions are shown below each sample.} 
  \label{fig:mae_images}
\end{figure}

\label{sup:additional_results_goal-image_prediction}
\boldparagraph{Goal-image prediction.}
We provide more qualitative examples for the goal-image prediction and the results during the training of masked auto-encoders, as shown in Fig.~\ref{fig:predictions} and Fig.~\ref{fig:mae_images}. In Fig.~\ref{fig:predictions} we show examples of the same input image but with different language instructions. The results show that our model can effectively predict different goal states given different language-based instructions and the initial observation. Namely, these results show that our model can interpret the input instructions, factorize different objects that need to be manipulated, and further understand the spatial location or direction in the scene. This indicates the learned visual-action representations after the self-supervised learning with the pretext task effectively associate visual states with action.
In Fig.~\ref{fig:mae_images}, we present example results in our pre-training phase. Our predicted goal images are blurry as in other MAE-based methods~\cite{guptaSiamese2023, weinzaepfel2022croco, he2022vlmae}. These results demonstrate that our approach successfully learns to predict the goal image.

\begin{figure}[t]
  \centering
  \includegraphics[width=1.0\linewidth]{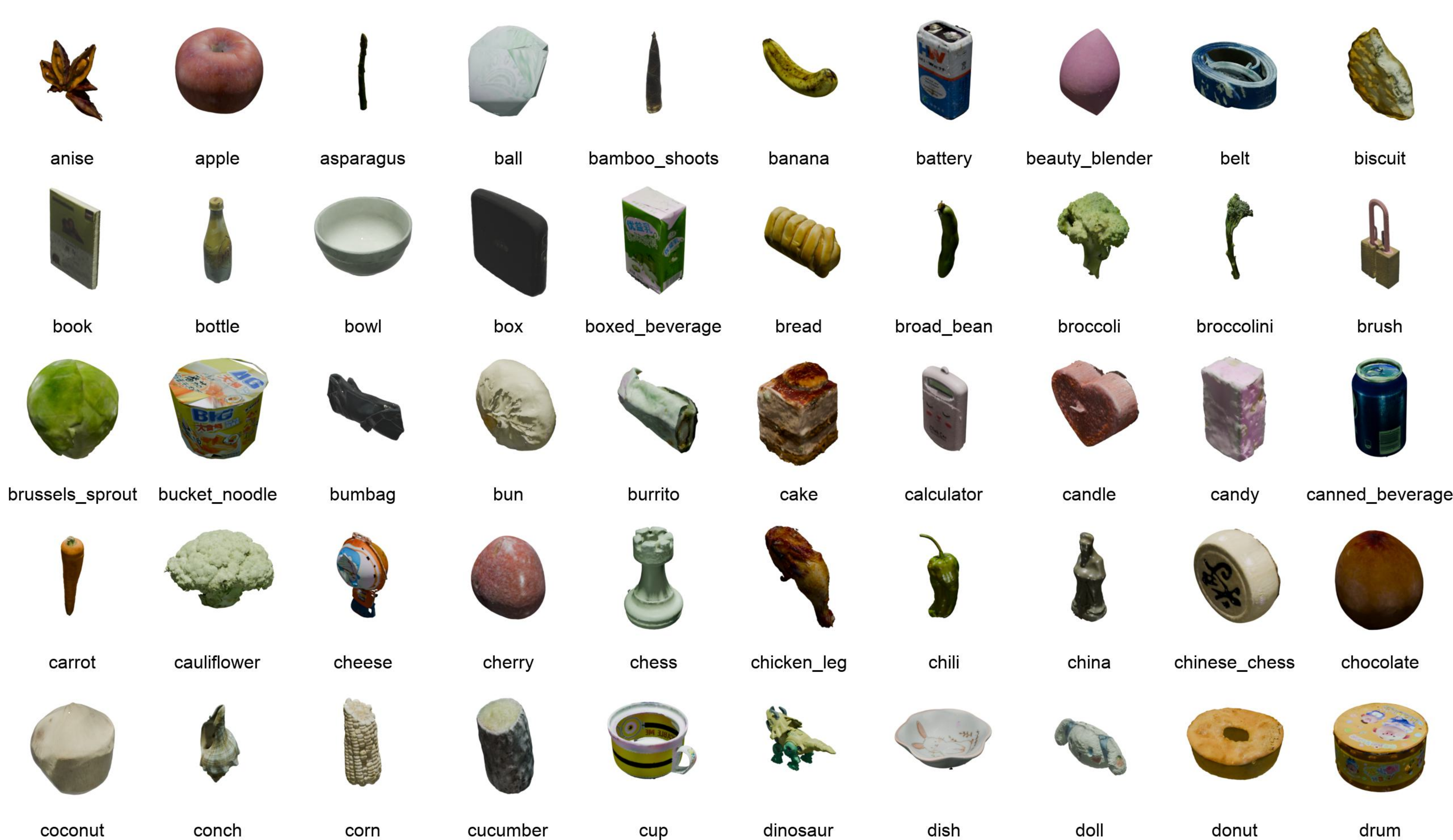}
 \caption{\textbf{Examples of objects from different classes.}} 
  \label{fig:objs1}
\end{figure}

\begin{figure}[b]
  \centering
  \includegraphics[width=1.0\linewidth]{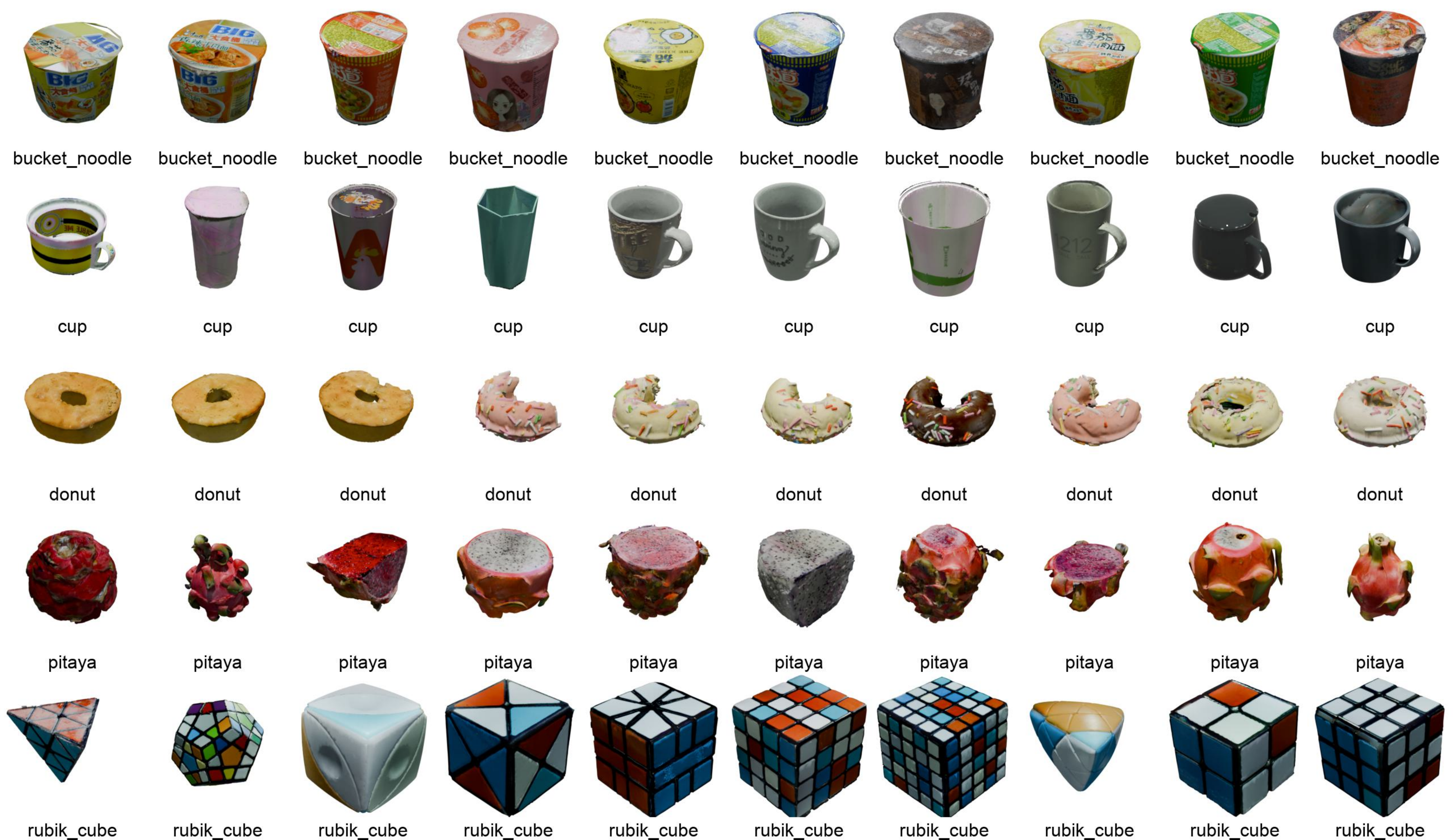}
 \caption{\textbf{Examples of intra-class variance.}} 
  \label{fig:objs2}
\end{figure}

\boldparagraph{Dataset details.} We build our Omni-Object Pick-and-Place (OOPP) dataset upon the previous benchmarsk~\cite{shridhar2022cliport, jiangVIMA2023} in the PyBullet Gym enviornment~\cite{benelot2018}. We manually selected 180 real-scanned object classes from the OmniObject3D dataset~\cite{omniobj}, focusing on those suitable for robotic manipulation, resulting in a total of 3,200 distinct instances. For each object, we reduced the mesh resolution to 20K faces to enable efficient rendering in simulation. Additionally, we filtered out objects that were either too large or too small (i.e., with dimensions less than 4cm or greater than 40cm). Our dataset includes full robot manipulation episodes, comprising of annotated robot actions, language instructions, and simulation-generated rewards. We use 160 object classes for training. For evaluating intra-class generalisation, we hold out a subset of instances from 20 categories included in the training set. For inter-class generalisation, we reserve 20 object categories that are entirely unseen during training. Fig.~\ref{fig:objs1} illustrates the first 50 examples across different object classes, while Fig.~\ref{fig:objs2} highlights examples of intra-class variation. The held-out categories for inter-class generalisation are sampled from four high-level semantic groups, as detailed in Table~\ref{tab:class_divisions}.

\begin{table}[htbp]
\centering
\caption{\textbf{Semantic group division of seen and unseen classes}. We divided all objects into four semantic groups. We selected unseen classes from each group in proportion to the number of object classes it contains, ensuring an even distribution of unseen categories.}
\label{tab:class_divisions}
\resizebox{\textwidth}{!}{%
\begin{tabular}{@{}ll>{\raggedright\arraybackslash}p{3.5cm}@{}}
\toprule
\textbf{Semantic Group} & \textbf{Seen Classes} & \textbf{Unseen Classes} \\ \midrule

\textbf{Food} &
\begin{tabular}[t]{@{}l@{}}
red\_jujube, corn, strawberry, anise, pizza, longan, loquat, chocolate, \\
brussels\_sprout, haw\_thorn, green\_bean\_cake, cucumber, litchi, cake, \\
dumpling, mooncake, rice\_cake, puff, water\_chestnut, mushroom, \\
broccoli, pastry, egg\_tart, kiwifruit, fig, cheese, chili, tomato, lemon, \\
oyster, steamed\_bun, carrot, mangosteen, bread, ginger, waffle, bun, \\
peach, apple, pear, potato, zongzi, pomegranate, onion, egg,\\
banana, chicken\_leg, sausage, coconut, broccolini, hami\_melon, \\
durian, asparagus, walnut, mango, loquat, bucket\_noodle
\end{tabular} &
orange, biscuit, shrimp, garlic, donut, sweet\_potato, candy, cherry, pancake \\ \midrule

\textbf{Daily-use} &
\begin{tabular}[t]{@{}l@{}}
thimble, beauty\_blender, battery, candle, calculator, plug, \\
watch, nipple, power\_strip, bottle, medicine\_bottle, tissue, \\
belt, dish, flash\_light, canned\_beverage, fork, cup, teapot, book,\\
glasses\_case, bowl, tape\_measure, speaker, laundry\_detergent,  \\
glasses, wallet, insole, bumbag, fan, knife, umbrella, kettle, light, \\
picnic\_basket, hammer, shoe, hat, laptop, vase, ornaments, spanner 
\end{tabular} &
soap, mouse, scissors, teapot, shampoo, toothpaste \\ \midrule

\textbf{Entertainment} &
\begin{tabular}[t]{@{}l@{}}
toy\_boat, toy\_plant, toy\_car, toy\_plane, timer, whistle, doll,\\
table\_tennis\_bat, toy\_motorcycle, drum, remote\_control, \\
garage\_kit, china, chess, Chinese\_chess, rubik\_cube, dinosaur
\end{tabular} &
toy\_bus, teddy\_bear, toy\_animals \\ \midrule

\textbf{Others} &
\begin{tabular}[t]{@{}l@{}}
hairpin, lotus\_root, house (model), plant, dumbbell, package, \\
bamboo\_shoots, brush, flute, ornaments, conch, magnet, box
\end{tabular} &
flower\_pot, red\_wine\_glass \\ \bottomrule
\end{tabular}%
}
\end{table}

\end{document}